\documentclass{ieeeaccess}
\usepackage{cite}
\usepackage{amsmath,amssymb,amsfonts}
\usepackage{algorithmic}
\usepackage{graphicx}
\usepackage{textcomp}
\usepackage{hyperref}
\usepackage{url}
\usepackage{bm}
\usepackage{tabularx}
\usepackage{float}
\usepackage{array}
\usepackage{placeins,color}
\usepackage[table,x11names,dvipsnames]{xcolor}
\usepackage[T1]{fontenc}
\usepackage{textcomp}
\usepackage{booktabs}
\usepackage{multirow}
\usepackage[utf8]{inputenc}
\usepackage{textgreek}
\usepackage{makecell} 
\usepackage{amsmath}
\makeatletter
\def\@IEEEbiographysepspace{\vspace{-5pt}}  
\makeatother

\AtBeginDocument{\DeclareMathVersion{bold}
\SetSymbolFont{operators}{bold}{T1}{times}{b}{n}
\SetSymbolFont{NewLetters}{bold}{T1}{times}{b}{it}
\SetMathAlphabet{\mathrm}{bold}{T1}{times}{b}{n}
\SetMathAlphabet{\mathit}{bold}{T1}{times}{b}{it}
\SetMathAlphabet{\mathbf}{bold}{T1}{times}{b}{n}
\SetMathAlphabet{\mathtt}{bold}{OT1}{pcr}{b}{n}
\SetSymbolFont{symbols}{bold}{OMS}{cmsy}{b}{n}
\renewcommand\boldmath{\@nomath\boldmath\mathversion{bold}}}
\makeatother

\def\BibTeX{{\rm B\kern-.05em{\sc i\kern-.025em b}\kern-.08em
    T\kern-.1667em\lower.7ex\hbox{E}\kern-.125emX}}
    
\newcommand{\ignore}[1]{}

\begin{document}
\history{Date of publication xxxx 00, 0000, date of current version xxxx 00, 0000.}
\doi{10.1109/ACCESS.2024.0429000}

\title{Deep Learning-Based Meat Freshness Detection with Segmentation and OOD-Aware Classification}
\author{\uppercase{Hutama Arif Bramantyo}\authorrefmark{1}, \uppercase{Mukarram Ali Faridi}\authorrefmark{1},
\uppercase{Rui Chen}\authorrefmark{2}, \uppercase{Clarissa Harris}\authorrefmark{2} and \uppercase{Yin Sun}\authorrefmark{1}}

\address[1]{Electrical \& Computer Engineering Department, College of Engineering, Auburn University, Auburn, AL 36849 USA}
\address[2]{Department of Agricultural and Environmental Sciences, College of Agriculture, Environment \& Nutrition Sciences, Tuskegee University, Tuskegee, AL 36088 USA}
\tfootnote{This research was supported in part by the NSF under grant no. CNS-2239677 and supported by USDA's the Agriculture Economics and Rural Communities Program (award no. 2023-69006-40213), Student Success and Workforce Development/1890 Universities Foundation (award no: 2021-38427-34837/FY22-SSWD-TU-34837), and Evans-Allen Program from the U.S. Department of Agriculture’s National Institute of Food and Agriculture.}

\markboth
{Bramantyo \headeretal: Deep Learning-Based Meat Freshness Detection with Segmentation and OOD-Aware Classification}
{Bramantyo \headeretal: Deep Learning-Based Meat Freshness Detection with Segmentation and OOD-Aware Classification}

\corresp{Corresponding author: Rui Chen (e-mail: rchen@tuskegee.edu).}

\begin{abstract}
In this study, we present a meat freshness classification framework from Red--Green--Blue (RGB) images that supports both packaged and unpackaged meat datasets. The system classifies four in-distribution (ID) meat classes and uses an out-of-distribution (OOD)-aware abstention mechanism that flags low-confidence samples as \textit{No Result}.
The pipeline combines U-Net-based segmentation with deep feature classifiers. Segmentation is used as a preprocessing step to isolate the meat region and reduce background, producing more consistent inputs for classification. The segmentation module achieved an Intersection over Union (IoU) of 75\% and a Dice coefficient of 82\%, producing standardized inputs for the classification stage. For classification, we benchmark five backbones: Residual Network-50 (ResNet-50), Vision Transformer-Base/16 (ViT-B/16), Swin Transformer-Tiny (Swin-T), EfficientNet-B0, and MobileNetV3-Small. We use nested $5\times3$ cross-validation (CV) for model selection and hyperparameter tuning. On the held-out ID test set, EfficientNet-B0 achieves the highest accuracy (98.10\%), followed by ResNet-50 and MobileNetV3-Small (both 97.63\%) and Swin-T (97.51\%), while ViT-B/16 is lower (94.42\%). We additionally evaluate OOD scoring and thresholding using standard OOD metrics and sensitivity analysis over the abstention threshold.
Finally, we report on-device latency using TensorFlow Lite (TFLite) on a smartphone, highlighting practical accuracy--latency trade-offs for future deployment.
\end{abstract}


\titlepgskip=-21pt

\maketitle

\section{Introduction}
\label{sec:introduction}

Rising meat consumption continues to increase the need for practical tools that help monitor food quality and safety~\cite{AlizadehSani2024}. Spoilage not only contributes to foodborne illness risk, but also causes economic loss and avoidable food waste~\cite{Pateiro2021,Xu2022}. In the United States, where meat consumption exceeds 224 pounds per capita annually~\cite{USDA2022}, fast and reliable screening of meat freshness can support both consumer confidence and operational decision-making.

Conventional freshness assessment methods, including chemical analysis and sensory evaluation~\cite{LIU2024111772,TAN200427,AMSA2015}, are often invasive, costly, and dependent on expert judgment~\cite{Mörlein_2019}. Although AMSA guidelines provide standardized visual criteria (for example, appearance and color cues)~\cite{mmb12473}, applying them consistently in high-throughput, real-world settings remains challenging. This motivates automated approaches that are objective, scalable, and suitable for everyday imaging conditions.

A range of sensor-based systems combined with machine learning, such as electronic noses, hyperspectral imaging, and gas sensor arrays, can provide precise freshness estimation~\cite{id7,id5,id6,id4}. However, their adoption is limited by specialized hardware requirements and controlled measurement conditions. In contrast, RGB-camera-based computer vision offers a low-cost and non-invasive alternative~\cite{zhou2019,id15,tahir2021survey}. Prior studies have shown that convolutional neural networks (CNNs) can capture color and texture cues associated with spoilage~\cite{ulucan2019,bhargav2023meat,abdelfattah2025,oforimensah2024}, sometimes reporting very high accuracy in controlled environments. In practice, however, performance can degrade due to background clutter, tray artifacts, glare, reflections, and packaging-related interference. These factors are especially common in packaged products.

Recent advances in Transformer-based vision models, such as the Vision Transformer (ViT) and Swin Transformer~\cite{dosovitskiy2021an,liu2021swin}, provide an alternative to CNN backbones by leveraging self-attention to model long-range dependencies. These architectures have shown strong results in challenging recognition tasks~\cite{id37,id39}. Yet, their use in meat freshness recognition remains limited, and few works study robustness under practical imaging issues such as glare, reflections, and packaging occlusions.

Overall, several gaps remain. First, much of the literature focuses on unpackaged meat~\cite{ulucan2019,bhargav2023meat,abdelfattah2025,oforimensah2024}, while packaged meat introduces additional visual complexity, including reflections, film glare, and label occlusions. Second, the impact of explicit segmentation on downstream classification is not consistently examined under a unified evaluation protocol, especially when comparing CNN and Transformer backbones. Third, most existing pipelines assume all test inputs belong to known classes and do not incorporate out-of-distribution (OOD) awareness~\cite{gal2016,hendrycks2017}. This can lead to over-confident predictions on unfamiliar conditions.

In this work, we address meat freshness recognition for both packaged and unpackaged images using a two-stage pipeline.
We first segment the meat region to reduce background and tray clutter, and then perform closed-set classification on four in-distribution (ID) meat classes.
To handle uncertain or potentially out-of-distribution inputs, we additionally incorporate an OOD-aware abstention mechanism: when confidence is low, the system returns \textit{No Result} rather than forcing a potentially overconfident prediction.
We evaluate CNN and Transformer backbones under a nested cross-validation protocol, and we report on-device latency using TensorFlow Lite to highlight deployment trade-offs. The main contributions are summarized below:

\begin{enumerate}
    \item 
    We study a unified RGB-based pipeline that handles both packaged and unpackaged meat images and includes an OOD-aware abstention option (\textit{No Result}) to avoid low-confidence decisions under challenging conditions.

    \item 
    We use U-Net-based segmentation to reduce background and tray clutter and to standardize the meat region prior to classification. We report segmentation agreement (IoU/Dice) under the same pseudo-labeling procedure used for training, and discuss how segmentation errors may affect downstream predictions.

    \item 
    We benchmark five backbones (ResNet-50, ViT-B/16, Swin-T, EfficientNet-B0, and MobileNetV3-Small) under a nested $5\times3$ cross-validation protocol for consistent model selection and hyperparameter tuning, enabling a direct accuracy--efficiency comparison across CNN and Transformer.

    \item 
    We evaluate OOD scoring and thresholding using standard OOD metrics and sensitivity analyses, and we measure on-device latency with TensorFlow Lite to report realistic deployment cost.
\end{enumerate}

The remainder of this paper is organized as follows: Section~\ref{sec:related_work} reviews related studies; Section~\ref{sec:methodology} describes the dataset and methodology; Section~\ref{sec:results} presents the experimental results and analysis; and Section~\ref{sec:conclusion} concludes the paper with future directions.

\begin{figure}[t]
  \centering
  \includegraphics[width=0.6\linewidth]{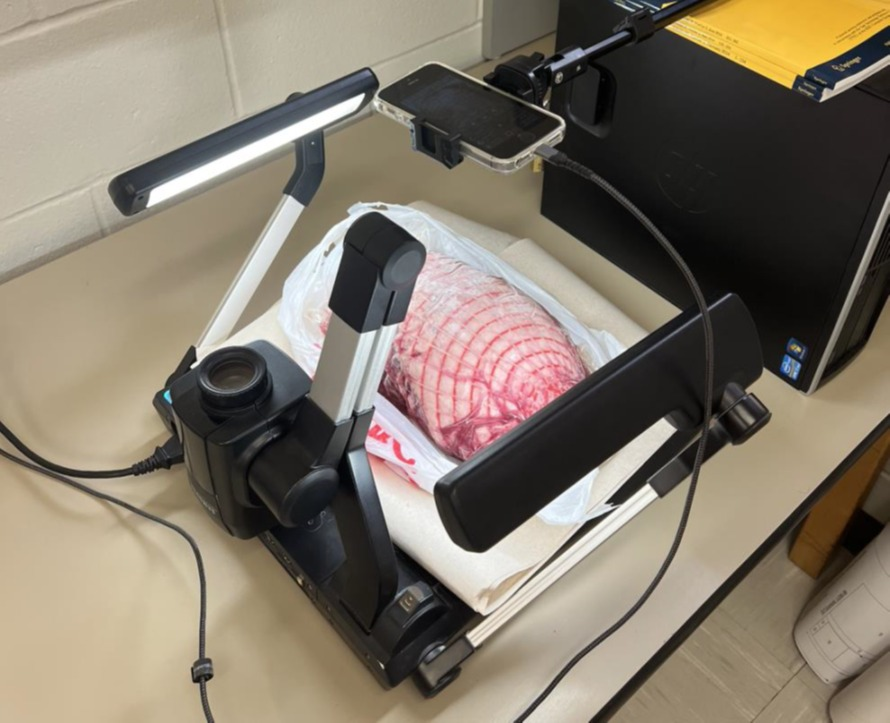}
  \caption{Experiment setup for packaged meat image collection.}
  \label{fig:experimentsetup}
\end{figure}

\section{Related Work}
\label{sec:related_work}

Early vision-based approaches to meat freshness assessment often relied on handcrafted features extracted from RGB images. More recent work has shifted toward convolutional neural networks (CNNs) to learn discriminative cues automatically. For example, Ulucan et al.~\cite{ulucan2019} reported high accuracy for binary classification of retail beef images under relatively controlled acquisition conditions. While the results are promising, many such studies focus on unpackaged meat and do not explicitly address practical issues such as glare, reflections, or packaging-related occlusions.

Several studies have extended freshness prediction beyond a binary setting. Bhargav et al.~\cite{bhargav2023meat} explored multi-class freshness recognition and introduced a cost-sensitive evaluation tailored to misclassification severity. However, prior work typically evaluates on limited datasets or controlled setups, and it rarely studies uncertainty handling for unfamiliar imaging conditions. Subsequent work has also explored deeper architectures and optimization strategies. Abdelfattah et al.~\cite{abdelfattah2025} combined a VGG19-based model with a meta-heuristic optimizer and reported strong performance for three-class prediction, while Ofori-Mensah et al.~\cite{oforimensah2024} compared CNN backbones for binary beef classification. Overall, these studies confirm that deep models can learn appearance cues associated with spoilage, but questions remain about generalization under real-world imaging variability and about mechanisms that make predictions safer when inputs fall outside the expected distribution.

Segmentation has been widely used as a preprocessing step in medical and natural-image analysis to isolate relevant regions, with U-Net being a common choice~\cite{ronneberger2015unet}. In food and agricultural imaging, lightweight models and detection-based pipelines have also been explored to support field deployment. For instance, MobileNet-based systems have been used for efficient visual diagnosis, and detection models such as YOLO have been applied to related tasks such as ripeness estimation~\cite{Karim2024,Majdudin2024}. These directions suggest that accuracy alone is not sufficient, and practical deployment constraints such as background clutter and inference-time cost should be considered.

A related but often overlooked aspect in meat freshness recognition is uncertainty and out-of-distribution (OOD) awareness. Confidence-based baselines such as maximum softmax probability (MSP) are commonly used~\cite{hendrycks2017}, and stronger scoring methods such as ODIN~\cite{liang2018odin} and energy-based scores~\cite{liu2020energy} have been proposed to improve OOD separability. These methods motivate evaluating not only in-distribution accuracy, but also how a system behaves when inputs are ambiguous or collected under unseen conditions.

Motivated by these gaps, our work studies both packaged and unpackaged meat images using a two-stage pipeline that combines segmentation and classification, and includes an OOD-aware abstention option. Concretely, the model predicts one of four meat classes when confident and returns \textit{No Result} when the input is uncertain or potentially out-of-distribution. We benchmark CNN, Transformer, and lightweight backbones under a nested cross-validation protocol and report on-device latency to ground the discussion in deployment constraints.

\section{Data and Methods}
\label{sec:methodology}
\subsection{Datasets}
\subsubsection{Data Collection}
To cover a broad range of freshness conditions, we combined several open-access datasets of unpackaged meat and collected a complementary packaged-meat dataset.
For unpackaged images, we used three open-access datasets:
\begin{itemize}
    \item Meat Quality Assessment Dataset (Izmir University)~\cite{id27}, containing 1,896 images evenly split between fresh and spoiled meat.
    \item Meat Freshness Image Dataset~\cite{id28}, which provides 1,641 fresh and 624 spoiled images. Images labeled as ``half-fresh'' were grouped with the fresh category.
    \item LOCBEEF Dataset~\cite{LOCBEEF2024}, comprising 3,268 images of Aceh beef (1,632 fresh, 1,636 spoiled) captured under various lighting and resolution settings to promote generalization.
\end{itemize}

We collected a new dataset in April 2024. 
Packaged meat products were sourced from local retailers in Auburn, Alabama (including Sam’s Club, Kroger, and Walmart). 
Then, we photographed the packaged meat in a controlled indoor environment using a smartphone mounted on a tripod, with temperature maintained at $25^{\circ}C$. 
Lighting intensity and direction were varied, including flash and ambient illumination, to reproduce realistic visual differences observed during spoilage. Differences across phone models and camera pipelines (e.g., auto white balance and exposure) are not explicitly evaluated here. 
To reduce dependence on a single setup, we varied illumination and viewing angles during capture, and we also used basic color/brightness augmentation during training.

The resulting dataset contained 3,734 packaged meat images (1,154 fresh, 2,580 spoiled). 
Figure~\ref{fig:experimentsetup} illustrates the indoor acquisition setup.

The in-distribution dataset was organized into four freshness categories:
(i) Unpackaged Fresh Meat, (ii) Unpackaged Spoiled Meat, (iii) Packaged Fresh Meat, and (iv) Packaged Spoiled Meat.
In addition, we curated a separate \textit{No Result} set for OOD/rejection evaluation.
Each packaged-meat image was labeled by its spoilage stage following AMSA visual assessment standards~\cite{mmb12473, AMSA_Meat_Judging}, which describe changes in surface color, moisture, and texture.

To maintain labeling consistency, we applied a chronological rule based on exposure time: images captured within 10--36 hours at room temperature ($25^{\circ}C$) were labeled as fresh, after which they were transitioned to the spoiled class. While we acknowledge that this duration is not a strict biochemical threshold, it serves as a practical visual freshness proxy. 
This protocol is motivated by general food-safety guidelines stating that detectable sensory changes in meat (color, odor, and texture) typically manifest within 24--48 hours under non-refrigerated conditions~\cite{FDA2023,Healthline2023,Jay2005,Orkusz2024}.

We note that this rule is not a biochemical threshold. We did not validate these labels using microbial measurements (e.g., total viable counts), and we did not run an expert sensory panel. 
AMSA is cited here as a visual reference for what changes to look for (color, surface moisture, texture), not as expert grading. 
Labeling was performed by student annotators using a short written guide with example images, and we did not measure inter-annotator agreement. 
Therefore, the packaged-meat labels should be interpreted as a visual freshness proxy under our protocol, and some label noise is expected near the transition period.

\begin{figure}[t] 
    \centering
    \includegraphics[width=0.8\linewidth]{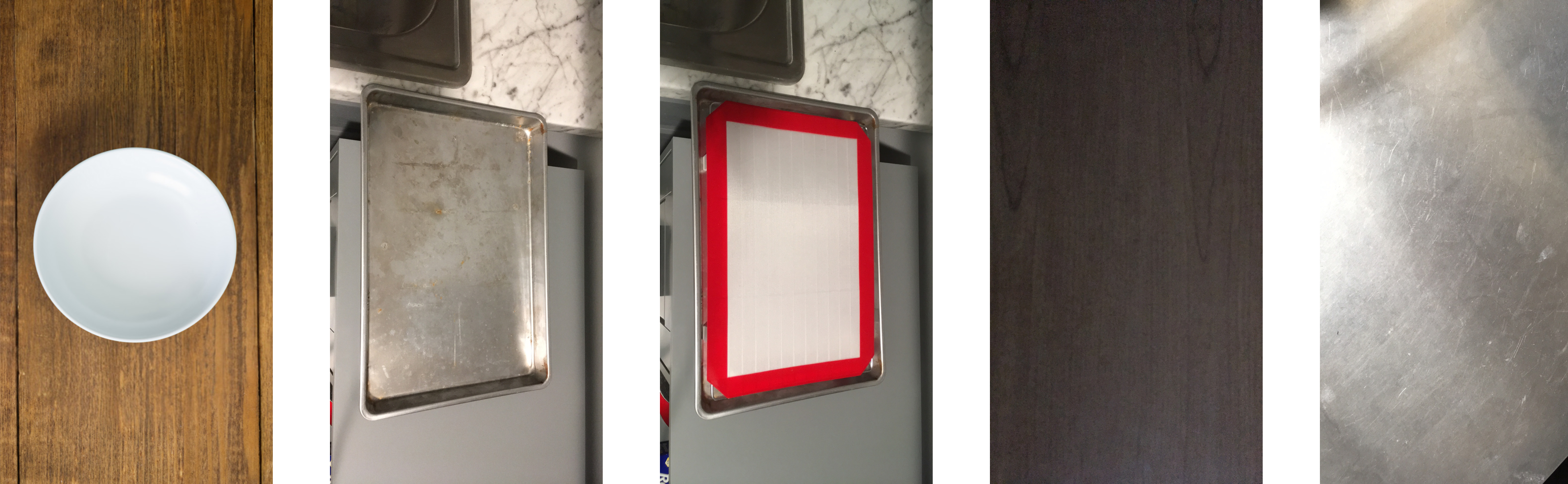}
    \caption{Representative examples of the \textit{No Result} (irrelevant/OOD) set, including empty trays, plates, and background-only images.}
    \label{fig:noresult}
\end{figure}

We also collected an additional set of irrelevant images (\textit{No Result}, 800 images, $\sim$13\%) such as empty trays, cutting boards, and background-only frames (Fig.~\ref{fig:noresult}). 
This set is used to evaluate the OOD/rejection mechanism (Section~\ref{sec_OOD}) and to emulate realistic failure cases where a camera captures non-meat scenes. 
It is kept separate from the four in-distribution freshness classes and is not used as a supervised class for freshness classification.

\subsubsection{Data Cleaning and Deduplication}

Before modeling, we audited the merged corpus of 11{,}163 images for duplicates and near-duplicates to mitigate the risk of split leakage and overly optimistic performance estimates \cite{BarzDenzler2020ciFAIR, Ramos_2025_ICCV}. We computed perceptual hashes (pHash) to flag near-identical frames primarily caused by small viewpoint or illumination shifts common in rapid-succession or burst-mode acquisition \cite{krawetz2010perceptual, buchner_imagehash}. If these near-duplicates were to persist across train/val/test splits or across folds in our nested cross-validation, the evaluation would be biased upward because the model would effectively be tested on scenes already encountered during training \cite{BarzDenzler2020ciFAIR, Ramos_2025_ICCV, WainerCawley2021}. Consequently, pHash filtering was strictly applied \emph{before} any data partitioning.

After removing 4{,}907 redundant frames (a 43.95\% reduction), the final dataset contained 6{,}256 unique images. To avoid over-pruning, we did not discard entire spoilage sequences; instead, filtering was restricted to near-duplicate clusters (defined by a small pHash distance), ensuring at least one representative image per capture condition was retained. We removed near-duplicate frames while keeping at least one representative per capture condition to reduce leakage risk. While this step may discard minor micro-variations, it significantly improves the statistical independence of the evaluation samples. As part of our future work, we plan to explore group-wise splitting, where each near-duplicate cluster is kept within a single fold to maximize variation while preventing leakage \cite{Joeres2025DataSAIL}.

\begin{table}[t]
    \centering
    \caption{Per-class distribution after cleaning and stratified 70/15/15 splitting for the four in-distribution freshness classes. Unpackaged sources: \cite{id27,id28,LOCBEEF2024}; Packaged: this study (Apr 2024).}
    \label{tab:dataset_summary}
    \begin{tabular}{lrrrrr}
        \toprule
        \textbf{Category} & \textbf{Train} & \textbf{Val} & \textbf{Test} & \textbf{Total} & \textbf{Share (\%)} \\
        \midrule
        Packaged Fresh Meat      & 656  & 138 & 146 & 940   & 15.0 \\
        Packaged Spoiled Meat    & 805  & 170 & 180 & 1{,}155 & 18.5 \\
        Unpackaged Fresh Meat    & 1{,}159 & 246 & 256 & 1{,}661 & 26.6 \\
        Unpackaged Spoiled Meat  & 1{,}187 & 252 & 261 & 1{,}700 & 27.2 \\
        \midrule
        \textbf{Total}           & 3{,}807 & 806 & 843 & \textbf{5{,}456} & 100.0 \\
        \bottomrule
    \end{tabular}
\end{table}

After deduplication, we retained 6,256 unique images in total, consisting of 5,456 in-distribution (ID) images across the four freshness classes (Table~\ref{tab:dataset_summary}) and an additional 800 \textit{No Result} images reserved exclusively for out-of-distribution (OOD) evaluation (Section~\ref{sec_OOD}). During training, we applied standard data augmentations (random rotation, horizontal flip, and color jitter) to improve robustness.

\subsection{Machine Learning Architectures and Evaluation Pipeline}

\subsubsection{Model Architectures}

We compared five backbones with different accuracy--efficiency trade-offs.
ResNet-50~\cite{7780459} is used as the CNN baseline.
We include two transformer models, ViT-B/16~\cite{dosovitskiy2021an} and Swin-T~\cite{liu2021swin}.
We also test two lightweight CNNs for mobile deployment, MobileNetV3-Small~\cite{howard2019searching} and EfficientNet-B0~\cite{tan2019efficientnet}.
All models were initialized with ImageNet-pretrained weights and trained with $224\times224$ inputs.

Hyperparameters were selected with the same nested cross-validation protocol (Section~\ref{sec:nestedcv}) for all five backbones.
MobileNetV3-Small and EfficientNet-B0 did not reuse the ResNet-50 configuration.
Each model was tuned independently in the inner loop using the same search space, and the best setting can differ by backbone.

To adapt each backbone to our four-class in-distribution task, we replaced the classifier with a small head: a fully connected layer (512 units, ReLU), dropout ($p=0.3$), and a softmax output.
Models were trained with categorical cross-entropy, where label smoothing was treated as a tunable hyperparameter (α $\in$ \{0.0, 0.1\}) selected in the inner loop of nested cross-validation (Table~\ref{tab:hyperparam_ranges}) and then applied in the final retraining using the best-performing configuration.

We fine-tuned the full networks rather than freezing the backbones.
We first trained the classification head, then unfroze backbone layers in stages with a smaller learning rate.
This helps stabilize training and reduces overfitting in visually similar classes~\cite{howard2018ulmfit,kornblith2019imagenet,mosbach2021stability}.

\subsubsection{U-Net Segmentation}

We use a U-Net segmentation model with ResNet-18/34/50 encoders to extract the meat region before classification. All models were initialized using ImageNet pretraining to leverage low-level visual features. In this setup, the encoder extracts hierarchical texture features, while the decoder recovers spatial details through skip connections, ensuring precise mask generation.

To train the segmentation model, we adopted a weakly supervised labeling strategy by generating pseudo--ground truth masks using an adaptive GrabCut procedure in automatic rectangle mode (five iterations)~\cite{rother2004grabcut,papandreou2015weakly,khoreva2017simple}. 
The initialization box covered approximately 90\% of the image and was perturbed by $\pm$10\% to robustly enclose the foreground, guided by color and texture cues in HSV and Lab spaces. 
We applied morphological opening and closing to suppress noise and smooth boundaries. 
To sanity-check pseudo-label quality, we manually reviewed 589 masks (approximately 11\% of the dataset) spanning both packaged and unpackaged images, confirming consistent separation of the meat region. 
Because these masks are automatically produced rather than expert annotations, IoU and Dice should be interpreted as agreement with the pseudo-labeling procedure rather than absolute segmentation accuracy. 
Our primary goal is background suppression (e.g., tray clutter and reflections) to help downstream classification focus on the meat interior region; a fully annotated segmentation benchmark is left for future work.

Hyperparameter tuning for each encoder was conducted independently through an exhaustive search of 108 candidate combinations, varying the batch size, learning rate (LR), weight decay (WD), loss formulation, and learning rate scheduler. Upon identifying the optimal configuration, the parameters were fixed for the final training stage to ensure reproducibility. All segmentation models were trained on $384 \times 384$ pixel images using the Adam optimizer, with early stopping based on validation loss to mitigate overfitting. 

Training was executed for up to 50 epochs using a hybrid objective function combining Binary Cross-Entropy and Dice loss to effectively balance region-level accuracy with boundary sharpness. Performance was quantified via Intersection-over-Union (IoU), Dice coefficient, precision, recall, and pixel accuracy, with 95\% confidence intervals (CI) derived from bootstrapping with 5,000 resamples. Finally, we zero out pixels outside the predicted mask to reduce background influence and encourage the classifier to rely on meat-region cues.

\subsubsection{Balancing and Loss Function}
To mitigate class imbalance among the four in-distribution freshness classes, we optionally used sampling-based rebalancing during training. When enabled, \texttt{WeightedRandom\allowbreak Sampler} upweighted under-represented classes so that mini-batches contained a more balanced mix of the four classes.

Models were trained with categorical cross-entropy (with label smoothing when specified in the training configuration). The \textit{No Result} images were not treated as a supervised class during training; they were held out and used only for OOD evaluation (Section~\ref{sec_OOD}).

\subsubsection{Nested Cross-Validation}
\label{sec:nestedcv}

We used a nested cross-validation (CV) design with 5 outer folds and 3 inner folds. 
Hyperparameters were selected exclusively on the inner folds, while the outer folds were used only for evaluation, reducing information leakage and providing a more reliable estimate of generalization.

To keep tuning computationally feasible, we adopted a two-stage inner-loop procedure aligned with transfer-learning practice. 
In \textit{Inner-1} (head warm-up), the backbone was frozen and we searched over the classification-head learning rate, weight decay, and label smoothing. 
In \textit{Inner-2} (unfreeze refinement), we unfroze the backbone and refined the top candidates by searching over the backbone learning rate and MixUp strength while keeping the best head configuration fixed. 
All models were trained with AdamW using separate parameter groups for the head and backbone, together with a warmup+decay epoch scheduler.
During the 1-epoch head warm-up, the backbone remained frozen and was then unfrozen for Inner-2 refinement.
All backbones (ResNet-50, Swin-T, ViT-B/16, EfficientNet-B0, and MobileNetV3-Small) followed the same tuning protocol and search space; the selected hyperparameters were determined independently per model. 
The best inner-loop configuration was then used for retraining within the corresponding outer fold, and the final retraining reused the most consistently selected configuration across outer folds.

Table~\ref{tab:hyperparam_ranges} summarizes the hyperparameter search space explored.

\begin{table}[t]
\centering
\caption{Hyperparameter search space explored during inner-loop tuning under the nested 5$\times$3 CV framework (classification).}
\label{tab:hyperparam_ranges}
\footnotesize
\renewcommand{\arraystretch}{1.05}
\setlength{\tabcolsep}{4pt}
\begin{tabular}{lc}
\hline
\textbf{Hyperparameter} & \textbf{Search Range / Setting} \\
\hline
Head learning rate (Inner-1)     & $1{\times}10^{-3}$--$3{\times}10^{-3}$ \\
Backbone learning rate (Inner-2) & $1{\times}10^{-5}$--$3{\times}10^{-4}$ \\
Weight decay (Inner-1)           & $1{\times}10^{-4}$--$1{\times}10^{-1}$ \\
Label smoothing $\alpha$         & $\{0.0,\,0.1\}$ \\
MixUp $\alpha$                   & $\{0.0,\,0.2\}$ \\
Batch size                       & 32 \\
Optimizer                        & AdamW \\
Scheduler                        & Warmup+decay \\
Head warm-up                     & 1 epoch \\
Warmup                           & 2 epochs \\
Early stopping                   & fixed per loop \\
\hline
\end{tabular}
\end{table}

\subsubsection{OOD-Aware Rejection Mechanism}\label{sec_OOD}
We add a simple rejection option based on prediction confidence to avoid forced low-confidence decisions.
Given an input image $x$, the classifier outputs softmax probabilities over the four freshness classes, and we use the maximum softmax probability (MSP) as a confidence score~\cite{hendrycks2017}:
$s(x)=\max_k p(y=k\mid x)$.
If $s(x)<\tau$, the model abstains from assigning a 4-class label; in deployment, such samples can be mapped to a separate \textit{No Result} output.

We additionally evaluate two widely used confidence scores: Energy and ODIN~\cite{liu2020energy,liang2018odin}.
Energy computes a score from the logits, while ODIN applies temperature scaling and a small input perturbation before computing confidence.
In this work, we distinguish \emph{semantic OOD} inputs from \emph{hard in-distribution} cases.
Semantic OOD refers to non-meat or irrelevant scenes that fall outside the four-class label space, and we operationalize this using the held-out \textit{No Result} set.
In contrast, artifacts such as glare, film reflections, occlusion, and borderline freshness are treated as hard in-distribution samples because they still contain meat and belong to one of the four target classes; the rejection mechanism is nevertheless useful for reducing overconfident predictions on these challenging inputs~\cite{guo2017calibration}.

To avoid relying on a single threshold choice, we sweep $\tau$ and report the resulting coverage--rejection trade-off (Figs.~\ref{fig:tau_cov}--\ref{fig:tau_rej}).
For a single reference operating point across models, we report results at $\tau=0.5$ as a simple baseline (not an optimum), since it lies near the transition where rejection begins to increase while coverage remains close to its maximum.
OOD evaluation uses the held-out \textit{No Result} set as semantic OOD examples, and we compare MSP, Energy, and ODIN under the same protocol in Section~\ref{sec:ood_detection}.

\subsubsection{Evaluation Metrics.}
We report two types of results: standard 4-class freshness classification performance and OOD detection performance using the held-out \textit{No Result} set.

For in-distribution classification, we compute categorical cross-entropy loss and macro-averaged precision, recall, and F1 over the four freshness classes. The loss is defined as
\begin{equation}
L = -\frac{1}{N} \sum_{i=1}^{N} \sum_{j=1}^{C} y_{i,j} \log(p_{i,j}),
\label{eq:cce}
\end{equation}
where \(N\) is the number of images, \(C=4\) is the number of in-distribution classes, \(y_{i,j}\) is the binary indicator of the true class, and \(p_{i,j}\) is the predicted probability.

\begin{table*}[t!]
\centering
\caption{Global segmentation performance of U-Net backbones with 95\% confidence intervals.}
\label{tab:unet_results}
\renewcommand{\arraystretch}{1.3}
\setlength{\tabcolsep}{3pt}
\scriptsize
\begin{tabular}{lccccc}
\toprule
Backbone & IoU & Dice & Precision & Recall & Pixel Acc. \\
\midrule
ResNet-18 & \makecell{0.7211 \\ \relax [0.703, 0.739]} & \makecell{0.8021 \\ \relax [0.786, 0.817]} & \makecell{0.7780 \\ \relax [0.761, 0.794]} & \makecell{0.8727 \\ \relax [0.857, 0.888]} & \makecell{0.8569 \\ \relax [0.846, 0.868]} \\
\midrule
ResNet-34 & \makecell{0.7494  \relax [0.731, 0.767]}& \makecell{0.8188  \relax [0.802, 0.835]}& \makecell{0.8186 \relax [0.802, 0.835]}& \makecell{0.8490 \\ \relax [0.832, 0.866]} & \makecell{0.8777  \relax [0.867, 0.889]}\\
\midrule
ResNet-50 & \makecell{0.7320 \\ \relax [0.713, 0.750]} & \makecell{0.8057 \\ \relax [0.789, 0.822]} & \makecell{0.8103 \\ \relax [0.793, 0.826]} & \makecell{0.8419 \\ \relax [0.825, 0.859]} & \makecell{0.8709 \\ \relax [0.860, 0.882]} \\
\bottomrule
\end{tabular}
\end{table*}

For OOD detection, we treat the four-class test split as in-distribution inputs and the \textit{No Result} set as OOD inputs. Using confidence scores from MSP, Energy, and ODIN (Section~\ref{sec_OOD}), we report AUROC, AUPR, and FPR@95TPR on this ID--OOD benchmark. We also run a threshold sensitivity sweep to show the coverage--rejection trade-off (Section~\ref{sec:ood_detection}, Figs.~\ref{fig:tau_cov}--\ref{fig:tau_rej}). For MSP/ODIN we sweep the confidence threshold \(\tau\), while for Energy we sweep score thresholds on the energy scale (details in Section~\ref{sec:ood_detection}).


\section{Results}
\label{sec:results}
\subsection{Segmentation Results}
\label{sec:results_seg}

Since the training masks were generated as pseudo--ground truth via GrabCut \cite{rother2004grabcut}, the segmentation metrics in this section should be interpreted as agreement with the automated labeling procedure rather than absolute accuracy against manual expert annotations \cite{papandreou2015weakly,khoreva2017simple}. Accordingly, we report these results to benchmark encoder consistency and to verify that the resulting masks are sufficiently reliable for foreground extraction prior to classification.

All segmentation metrics are reported on the held-out ID test split (the same split used for downstream classification) to assess generalization on unseen images. These test-set metrics are used for evaluation only; model selection and early stopping were performed using the training/validation splits.

Table~\ref{tab:unet_results} summarizes the global performance across three ResNet backbones. ResNet-34 achieved the highest mean IoU (0.7494), Dice (0.8188), precision (0.8186), and pixel accuracy (0.8777). While ResNet-18 exhibited the highest recall (0.8727), minimizing missed meat pixels at the expense of higher false positives, ResNet-50 trailed in overall overlap metrics. Per-class IoU analysis (Table~\ref{tab:unet_perclass}) confirms this trend, with ResNet-34 leading in three out of four categories. Given this robust balance across overlap and precision, ResNet-34 was selected as the fixed encoder for the subsequent classification pipeline.

Integrating U-Net segmentation standardized the region of interest (ROI) and reduced background variation (e.g., tray edges and plastic reflections), supporting more stable feature learning \cite{ronneberger2015unet}. Because our objective is background suppression rather than pixel-perfect boundaries, minor pseudo-label boundary noise is unlikely to affect classification as long as the meat interior color and texture cues are preserved. We therefore treat segmentation as a preprocessing assumption of the pipeline; a dedicated ablation study (same backbone and evaluation protocol) comparing classification performance with and without segmentation is left for future work. To summarize stability for the selected ResNet-34 encoder, image-level bootstrapping yielded a Dice of 0.8188 with a 95\% CI of [0.802, 0.835], supporting consistent foreground extraction for the classifiers evaluated in Section~\ref{sec:results_nestedcv}.

\begin{table}[t]
\centering
\caption{Per-class IoU performance with 95\% confidence intervals.}
\label{tab:unet_perclass}
\renewcommand{\arraystretch}{1.3}
\setlength{\tabcolsep}{3.5pt}
\scriptsize
\begin{tabular}{lcccc}
\toprule
Backbone & \makecell{Packaged\\Fresh} & \makecell{Packaged\\Spoiled} & \makecell{Unpackaged\\Fresh} & \makecell{Unpackaged\\Spoiled} \\
\midrule
ResNet-18 & \makecell{0.8726 \\ \relax [0.842, 0.900]} & \makecell{0.7749 \\ \relax [0.741, 0.808]} & \makecell{0.6693 \\ \relax [0.639, 0.699]} & \makecell{0.6501 \\ \relax [0.612, 0.687]} \\
\midrule
ResNet-34 & \makecell{0.8888 \\ \relax [0.859, 0.915]} & \makecell{0.8072 \\ \relax [0.771, 0.842]} & \makecell{0.6898 \\ \relax [0.659, 0.722]} & \makecell{0.6900 \\ \relax [0.653, 0.728]} \\
\midrule
ResNet-50 & \makecell{0.8900 \\ \relax [0.858, 0.918]} & \makecell{0.7997 \\ \relax [0.763, 0.834]} & \makecell{0.6738 \\ \relax [0.643, 0.704]} & \makecell{0.6541 \\ \relax [0.615, 0.693]} \\
\bottomrule
\end{tabular}
\end{table}

\subsection{Nested Cross-Validation Performance}
\label{sec:results_nestedcv}

\begin{table}[t]
\centering
\caption{Nested CV stability (outer-fold mean$\pm$SD).}
\label{tab:nestedcv_stability}
\renewcommand{\arraystretch}{1.15}
\setlength{\tabcolsep}{4.5pt}
\scriptsize
\begin{tabular}{lcccc}
\toprule
Model & Acc (\%) & F1 (\%) & Prec. (\%) & Rec. (\%) \\
\midrule
ResNet-50        & 98.47$\pm$0.37 & 98.47$\pm$0.37 & 98.51$\pm$0.35 & 98.47$\pm$0.37 \\
Swin-T           & 97.75$\pm$0.64 & 97.75$\pm$0.64 & 97.80$\pm$0.62 & 97.75$\pm$0.64 \\
ViT-B/16         & 94.16$\pm$1.03 & 94.13$\pm$1.05 & 94.67$\pm$0.76 & 94.16$\pm$1.03 \\
EfficientNet-B0  & 97.88$\pm$0.72 & 97.88$\pm$0.72 & 97.96$\pm$0.67 & 97.88$\pm$0.72 \\
MobileNetV3-Small& 96.64$\pm$0.93 & 96.63$\pm$0.93 & 96.91$\pm$0.79 & 96.64$\pm$0.93 \\
\bottomrule
\end{tabular}
\end{table}

\begin{table*}[t!]
\centering
\caption{Nested cross-validation (5 outer $\times$ 3 inner) results and best hyperparameters per model.}
\label{tab:nested_summary}
\renewcommand{\arraystretch}{1.15}
\begin{tabular}{lccccccc}
\hline
\textbf{Model} & \textbf{Acc. (Mean $\pm$ SD, \%)} & \textbf{Head LR} & \textbf{Backbone LR} & \textbf{Weight Decay} & \textbf{Label Smooth} & \textbf{Mixup $\alpha$} & \textbf{Best Epoch} \\
\hline
ResNet-50        & 98.47 $\pm$ 0.37 & 0.003  & 1e$-$4  & 0.0005 & 0.0 & 0.0 & 19--30 \\
Swin-T           & 97.75 $\pm$ 0.64 & 0.0015 & 7e$-$5  & 0.1    & 0.0 & 0.0 & 13--30 \\
ViT-B/16         & 94.16 $\pm$ 1.03 & 0.001  & 5e$-$5  & 0.05   & 0.1 & 0.0 & 13--30 \\
EfficientNet-B0  & 97.88 $\pm$ 0.72 & 0.001  & 0.0003  & 0.0001 & 0.1 & 0.0 & 17--30 \\
MobileNetV3-Small& 96.64 $\pm$ 0.93 & 0.001  & 0.0003  & 0.0005 & 0.0 & 0.2 & 17--30 \\
\hline
\end{tabular}
\end{table*}

\begin{table*}[t]
\centering
\caption{Outer-fold accuracies (\%) under nested 5$\times$3 CV.}
\label{tab:supp_nestedcv_perfold}
\renewcommand{\arraystretch}{1.15}
\begin{tabular}{lccccc}
\hline
\textbf{Outer Fold} &
\textbf{ResNet-50 (Acc \%)} &
\textbf{Swin-T (Acc \%)} &
\textbf{ViT-B/16 (Acc \%)} &
\textbf{EfficientNet-B0 (Acc \%)} &
\textbf{MobileNetV3-Small (Acc \%)} \\
\hline
Fold 1 & 98.12 & 97.18 & 93.05 & 97.92 & 95.09 \\
Fold 2 & 98.36 & 97.61 & 94.92 & 97.64 & 97.26 \\
Fold 3 & 98.89 & 98.24 & 95.21 & 98.11 & 96.69 \\
Fold 4 & 98.53 & 97.82 & 93.84 & 96.88 & 97.45 \\
Fold 5 & 98.45 & 97.89 & 93.76 & 98.87 & 96.69 \\
\hline
\textbf{Mean $\pm$ SD} &
\textbf{98.47 $\pm$ 0.37} &
\textbf{97.75 $\pm$ 0.64} &
\textbf{94.16 $\pm$ 1.03} &
\textbf{97.88 $\pm$ 0.72} &
\textbf{96.64 $\pm$ 0.93} \\
\hline
\end{tabular}
\vspace{2mm}
\end{table*}

\begin{table}[t]
\centering
\caption{Retrain results on the held-out ID test set.}
\label{tab:retrain_all}
\scriptsize
\renewcommand{\arraystretch}{1.1}
\setlength{\tabcolsep}{3.5pt}
\begin{tabular}{lccc}
\hline
Model & Train & Val & Test (ID) \\
\hline
ResNet-50           & 99.9 & 98.5 & 97.63 \\
Swin-T              & 99.7 & 98.6 & 97.51 \\
ViT-B/16            & 99.1 & 94.2 & 94.42 \\
EfficientNet-B0     & 99.9 & 97.2 & 98.10 \\
MobileNetV3-Small   & 99.9 & 97.5 & 97.63 \\
\hline
\end{tabular}
\end{table}

All classifiers were evaluated with a nested $5\times3$ cross-validation protocol. 
The inner loop performed hyperparameter tuning and the outer loop measured generalization, 
thereby preventing information leakage and keeping estimates unbiased. 
Table~\ref{tab:nestedcv_stability} reports outer-fold mean$\pm$SD for accuracy, macro-F1, precision, and recall; 
Table~\ref{tab:nested_summary} summarizes the selected hyperparameters.

ResNet-50 obtained the highest outer-fold accuracy ($98.47\%\pm0.37$), 
followed by Swin-T ($97.75\%\pm0.64$) and ViT-B/16 ($94.16\%\pm1.03$). 
Convolutional inductive bias and Swin’s shifted-window hierarchy likely aid stable training on this dataset, 
whereas ViTs typically require larger and more diverse data with stronger regularization to capture spatial structure~\cite{dosovitskiy2021an,touvron2021deit}.

Per-fold outer-loop accuracies (Table~\ref{tab:supp_nestedcv_perfold}) show tight clustering across folds. 
For ResNet-50 the five outer folds span 98.12–98.89\% (range 0.77\%, SD 0.37\%), 
for Swin-T 97.18–98.24\% (range 1.06\%, SD 0.64\%), 
and for ViT-B/16 93.05–95.21\% (range 2.16\%, SD 1.03\%). 
This pattern matches the macro-averaged summary in Table~\ref{tab:nestedcv_stability} and indicates low dispersion without fold-specific drift. 
Best epochs also fell within a narrow band for each model (ResNet-50: 19–30, Swin-T: 13–30, ViT-B/16: 13–30), 
and the inner loop repeatedly selected similar learning-rate and regularization settings (Table~\ref{tab:nested_summary}), 
both of which are consistent with stable optimization across folds.

The hyperparameters in Table~\ref{tab:nested_summary} are the configurations most frequently selected by the inner loop and were reused for the final retraining with OOD-enabled inference to keep test results independent of tuning.

\begin{figure*}[!t]
\centering
\begin{tabular}{ccc}
\includegraphics[width=0.32\textwidth]{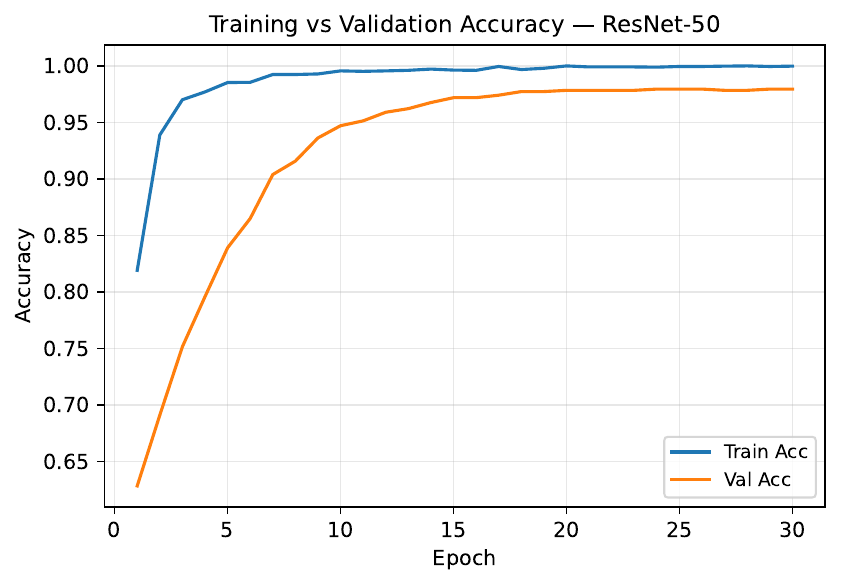} &
\includegraphics[width=0.32\textwidth]{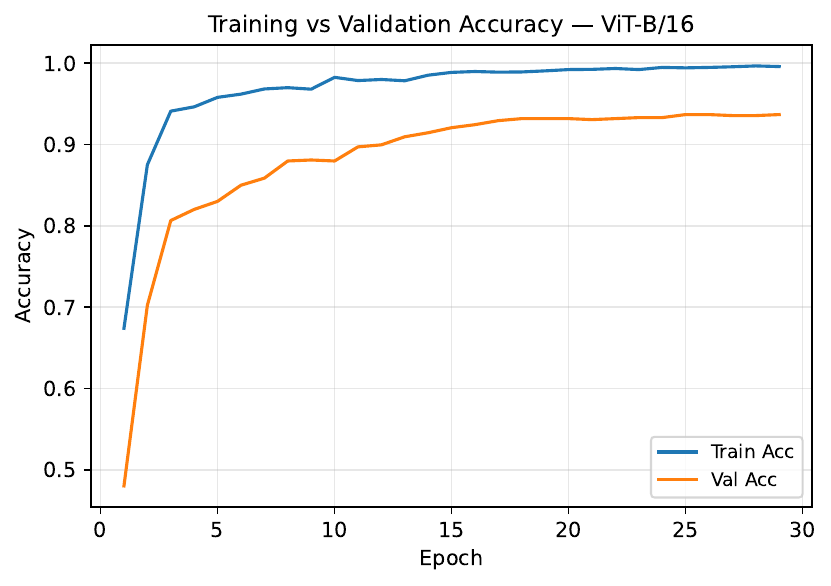} &
\includegraphics[width=0.32\textwidth]{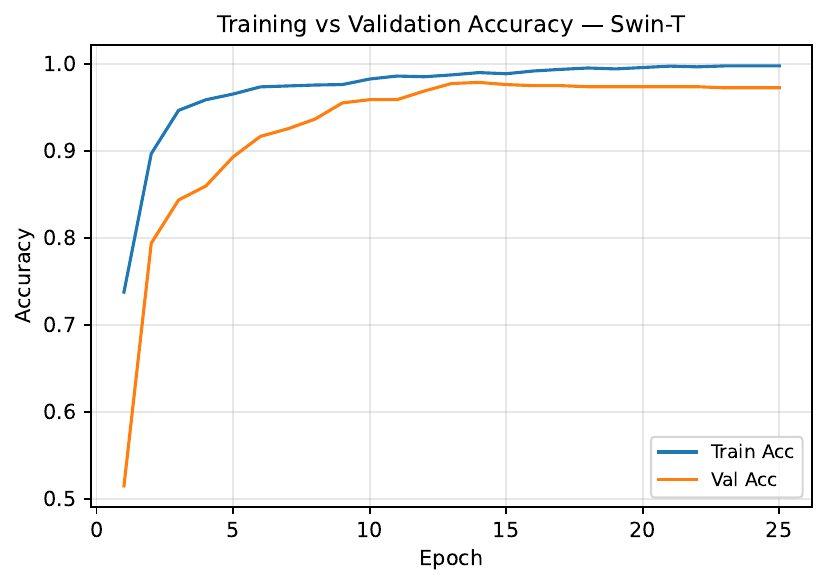} \\
(a) ResNet-50 & (b) ViT-B/16 & (c) Swin-T \\
\\[1mm]

\includegraphics[width=0.32\textwidth]{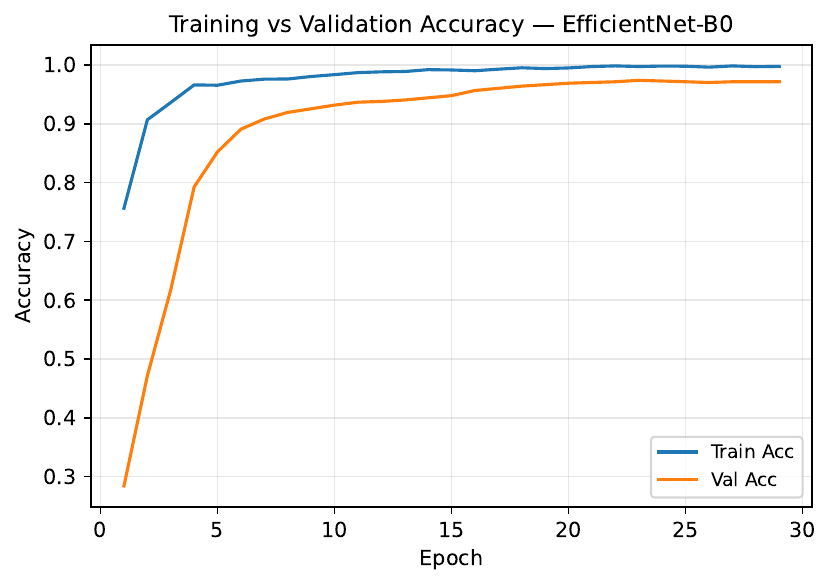} &
\includegraphics[width=0.32\textwidth]{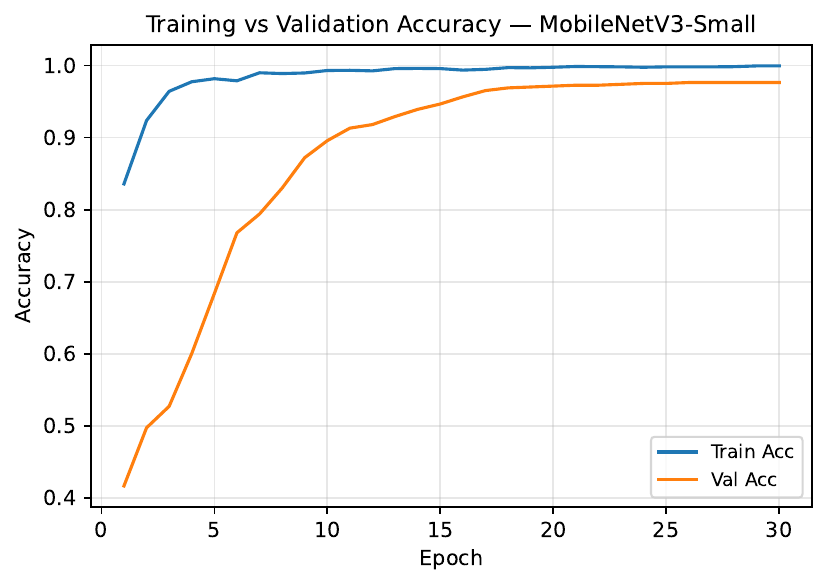} &
\rule{0.32\textwidth}{0pt} \\  
(d) EfficientNet-B0 & (e) MobileNetV3-Small & \\
\end{tabular}
\caption{Training vs validation accuracy curves for five models during retraining with OOD-enabled inference.}
\label{fig:trainval_all}
\end{figure*}

\subsection{Retrain \& Testing Performance}
\label{sec:retrain_ood}
Figures~\ref{fig:trainval_all}(a)--(e) show the training and validation accuracy curves during the final retraining stage on the four in-distribution (ID) classes.
All backbones converge stably, with ResNet-50 and Swin-T exhibiting smooth training dynamics and a small train--val gap.
EfficientNet-B0 and MobileNetV3-Small also converge quickly, but show a slightly larger separation between training and validation.
ViT-B/16 improves more gradually and exhibits the largest train--val gap, which is consistent with its lower validation accuracy (94.2\%) and the lowest held-out ID test accuracy in Table~\ref{tab:retrain_all}.

Table~\ref{tab:retrain_all} summarizes the final retraining and held-out ID test results ($N=843$).
EfficientNet-B0 achieves the highest ID test accuracy (98.10\%), followed by ResNet-50 and MobileNetV3-Small (both 97.63\%), and Swin-T (97.51\%).
ViT-B/16 is lower (94.42\%), aligning with its reduced validation accuracy (94.2\%) during retraining.
Overall, the validation ranking broadly matches the test trend, suggesting that the selected backbones generalize reliably to the held-out ID set.

\begin{table*}[!t]
\centering
\caption{Per-class performance of five models on the held-out ID test set ($N=843$).}
\label{tab:perclass_models}
\renewcommand{\arraystretch}{1.15}
\setlength{\tabcolsep}{3.2pt}
\scriptsize
\begin{tabular}{lccccccccccccccc}
\toprule
\multirow{2}{*}{Class} &
\multicolumn{3}{c}{ResNet-50} &
\multicolumn{3}{c}{Swin-T} &
\multicolumn{3}{c}{ViT-B/16} &
\multicolumn{3}{c}{EfficientNet-B0} &
\multicolumn{3}{c}{MobileNetV3-Small} \\
\cline{2-16}
& Prec. & Rec. & F1
& Prec. & Rec. & F1
& Prec. & Rec. & F1
& Prec. & Rec. & F1
& Prec. & Rec. & F1 \\
\midrule
Packaged Fresh     & 0.986 & 0.952 & 0.969 & 0.935 & 0.993 & 0.963 & 0.971 & 0.911 & 0.940 & 0.973 & 0.993 & 0.983 & 0.954 & 0.986 & 0.970 \\
Packaged Spoiled   & 0.962 & 0.989 & 0.975 & 0.994 & 0.944 & 0.969 & 0.931 & 0.978 & 0.954 & 0.994 & 0.978 & 0.986 & 0.989 & 0.961 & 0.975 \\
Unpackaged Fresh   & 0.959 & 1.000 & 0.979 & 0.992 & 0.969 & 0.980 & 0.895 & 1.000 & 0.945 & 0.959 & 1.000 & 0.979 & 0.959 & 1.000 & 0.979 \\
Unpackaged Spoiled & 1.000 & 0.958 & 0.978 & 0.970 & 0.992 & 0.981 & 1.000 & 0.885 & 0.939 & 1.000 & 0.958 & 0.978 & 1.000 & 0.958 & 0.978 \\
\midrule
Macro Avg          & 0.977 & 0.975 & 0.975 & 0.973 & 0.975 & 0.973 & 0.949 & 0.943 & 0.944 & 0.982 & 0.982 & 0.982 & 0.975 & 0.976 & 0.975 \\
\bottomrule
\end{tabular}
\end{table*}

\begin{figure*}[t]
\centering
\includegraphics[width=0.32\linewidth]{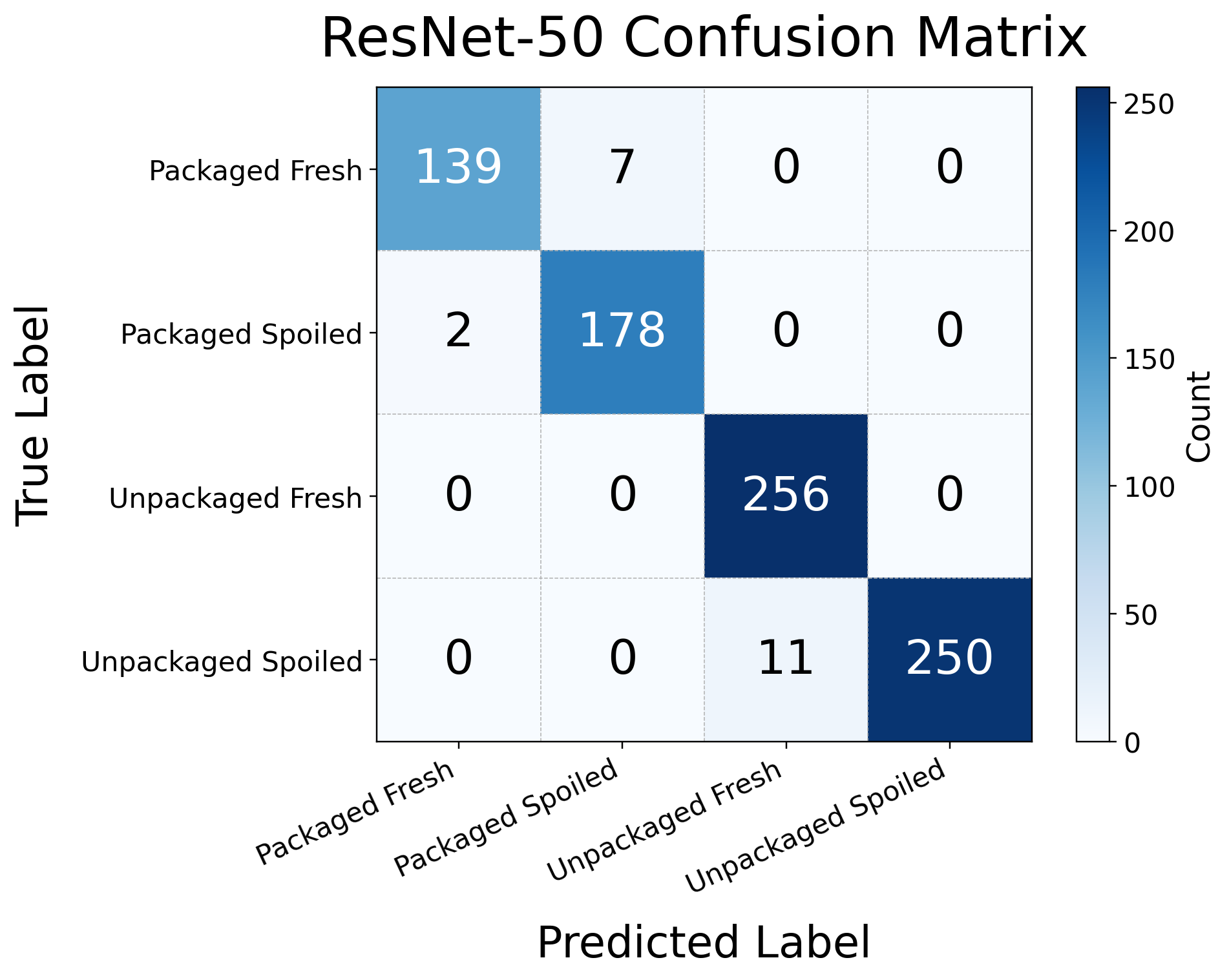}\hfill
\includegraphics[width=0.32\linewidth]{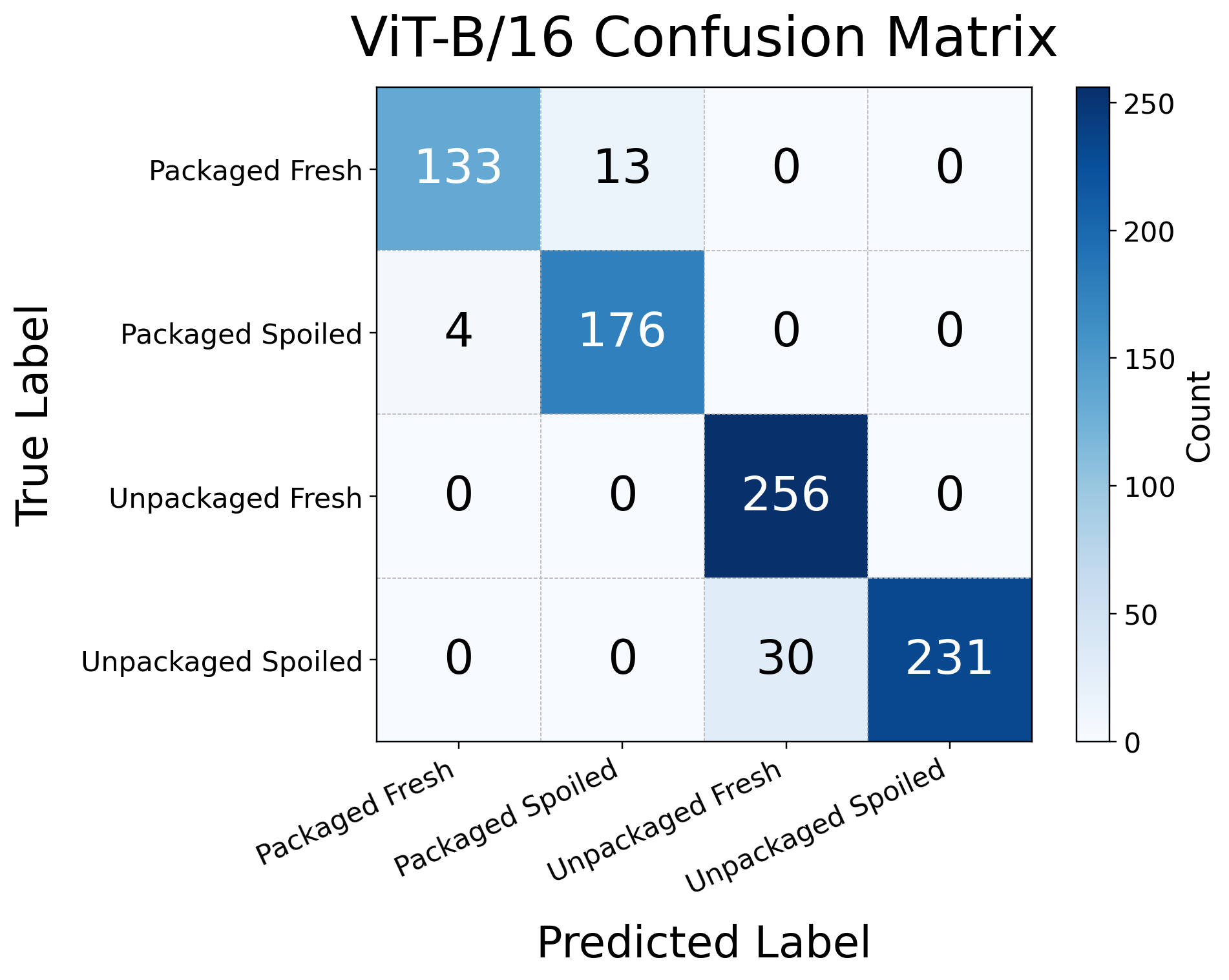}\hfill
\includegraphics[width=0.32\linewidth]{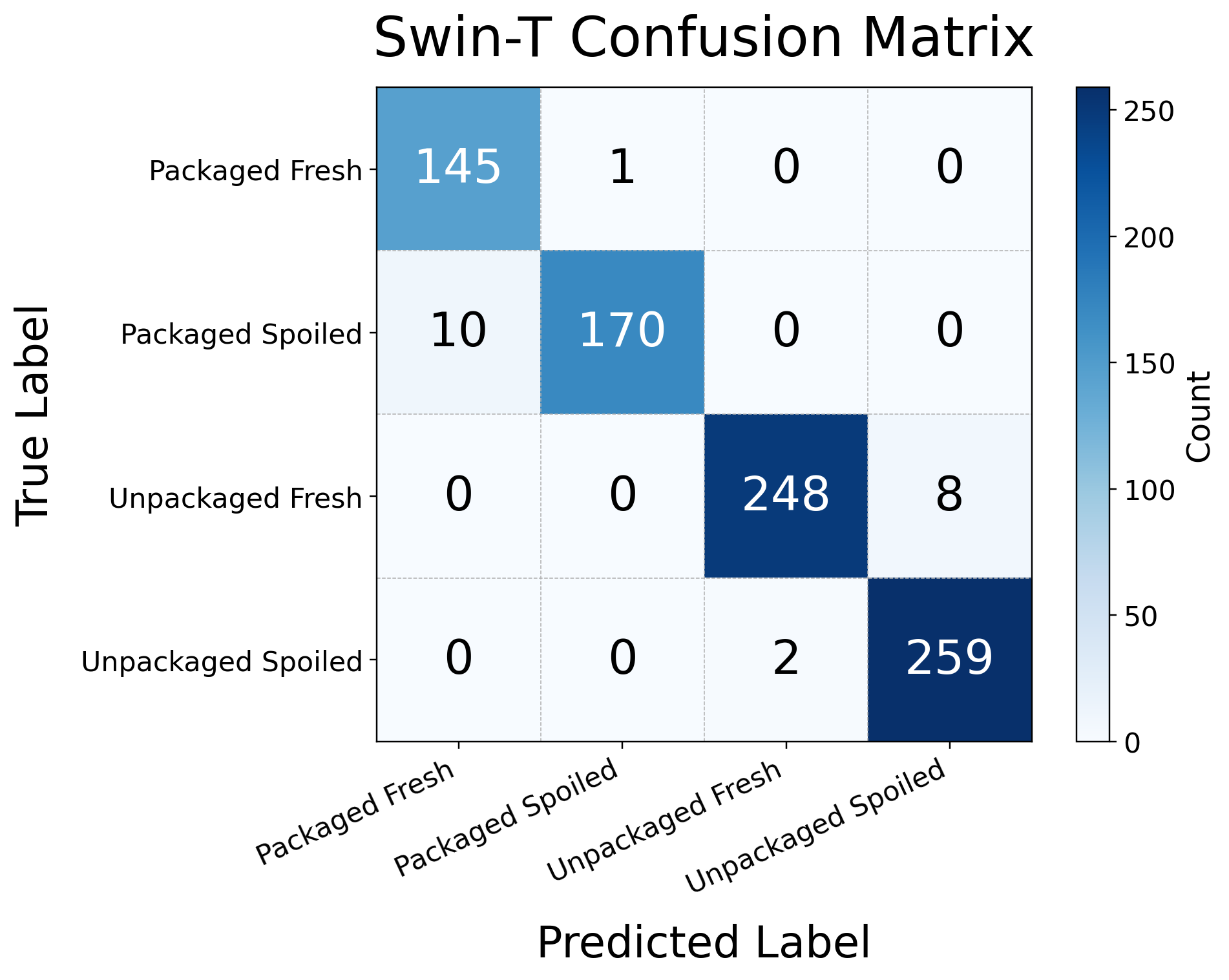}\\[4pt]
\includegraphics[width=0.32\linewidth]{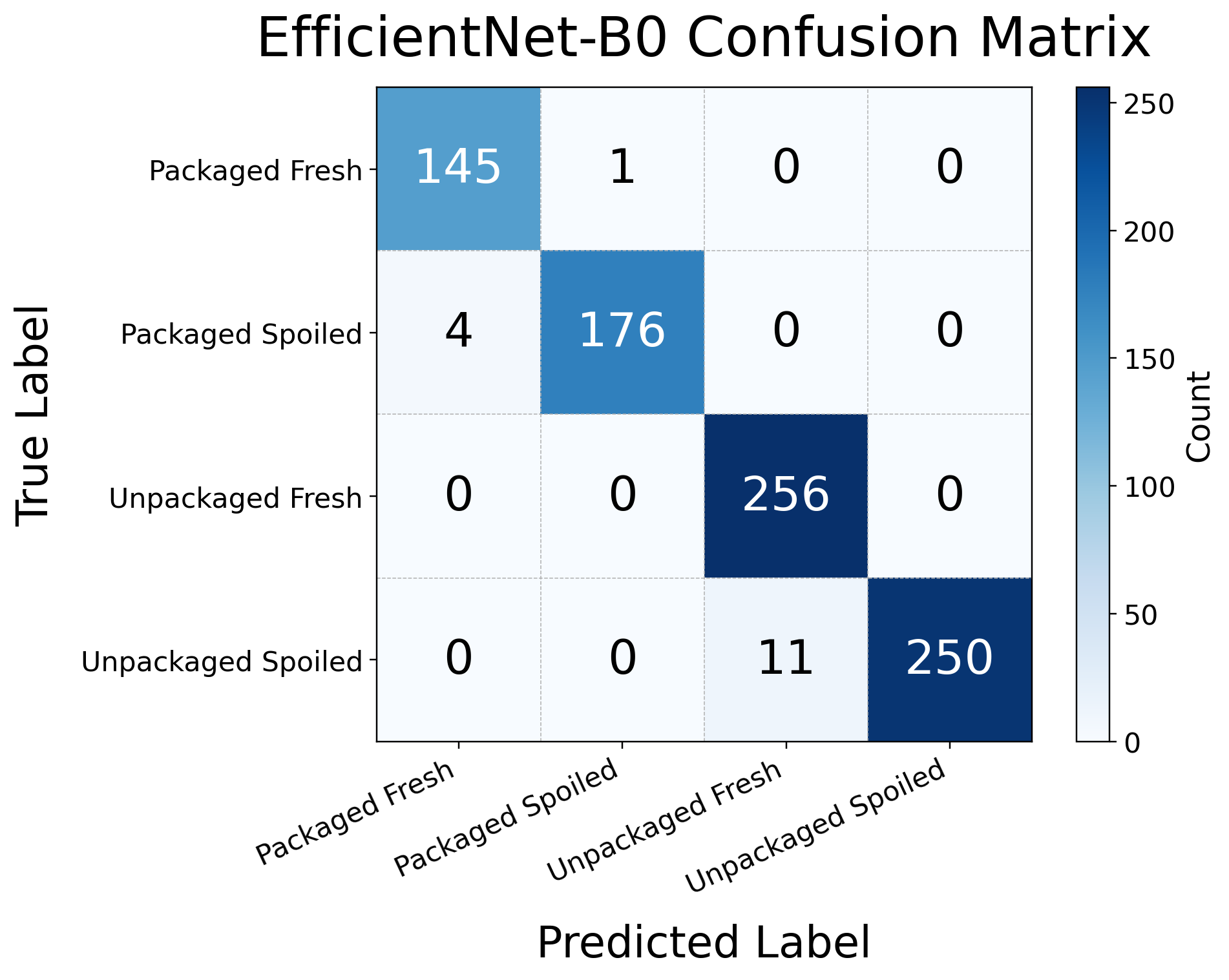}\hfill
\includegraphics[width=0.32\linewidth]{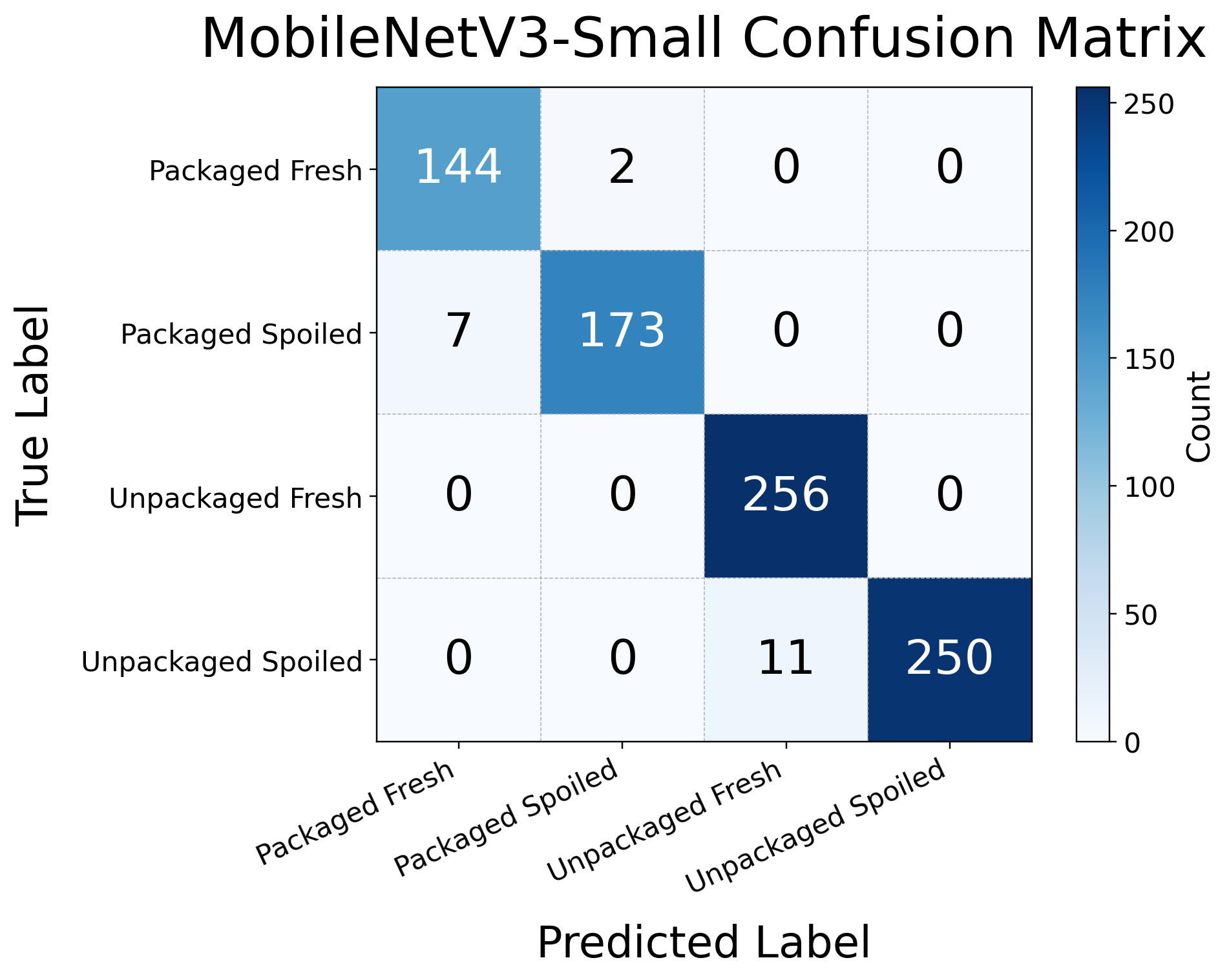}
\caption{Confusion matrices on the held-out ID test set ($N=843$) for the four in-distribution classes (Packaged Fresh, Packaged Spoiled, Unpackaged Fresh, Unpackaged Spoiled). Errors mostly occur within the same packaging type (fresh vs.\ spoiled), which is the hardest visual boundary.}
\label{fig:cm_all_models}
\end{figure*}

Per-class results in Table~\ref{tab:perclass_models} show consistent strengths and weaknesses across backbones.
EfficientNet-B0 achieves the best macro-average F1 (0.982), while ResNet-50 and MobileNetV3-Small are close behind (both 0.975), with Swin-T at 0.973.
ViT-B/16 is lower (0.944), with the largest drops appearing in the packaged classes (e.g., Packaged Fresh recall 0.911 and Unpackaged Spoiled recall 0.885), which are more sensitive to glare, film reflections, and subtle appearance changes.
Across all models, unpackaged classes remain comparatively easier, with near-saturated recall for Unpackaged Fresh in several backbones.

\begin{figure*}[t]
\centering
\includegraphics[width=0.31\textwidth]{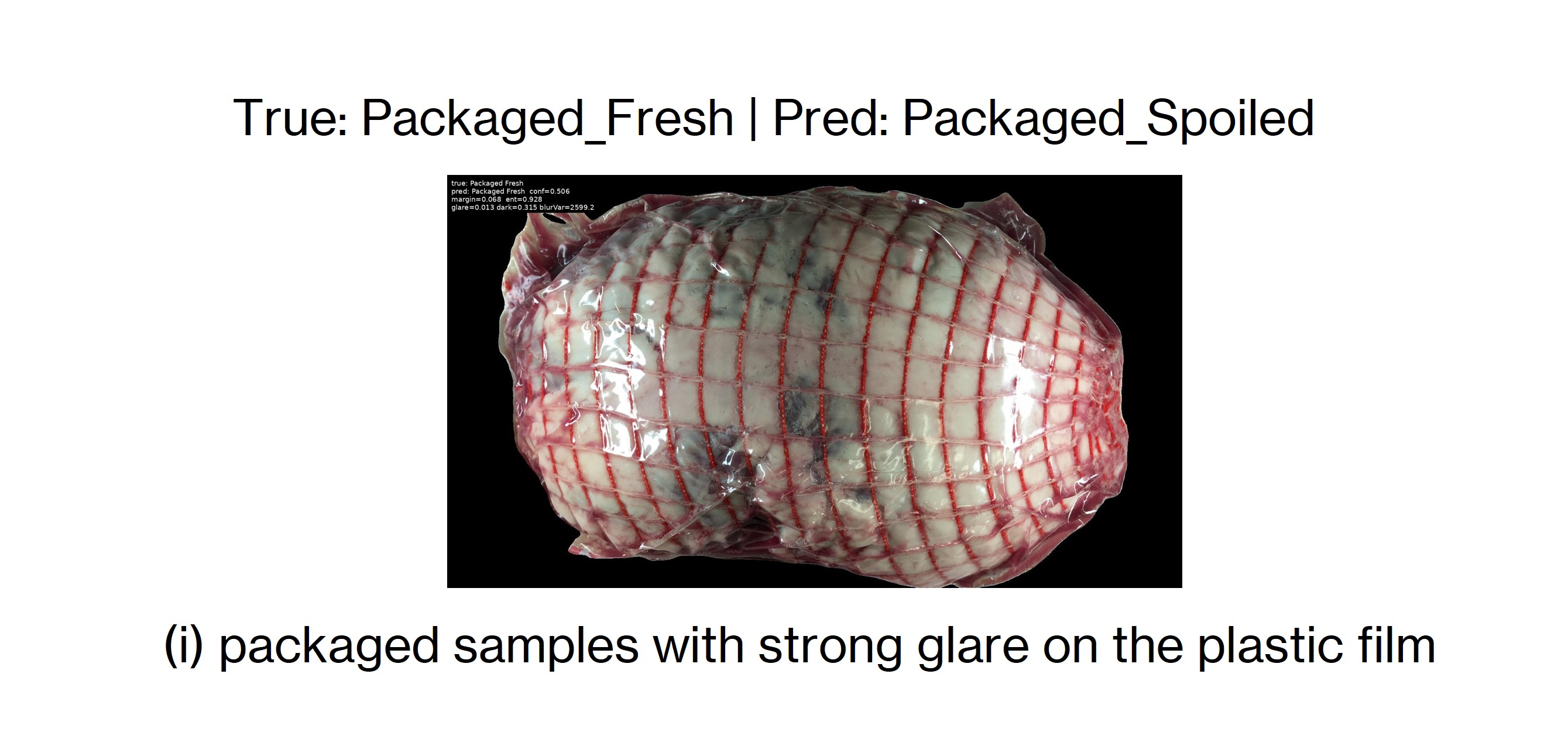}\hfill
\includegraphics[width=0.31\textwidth]{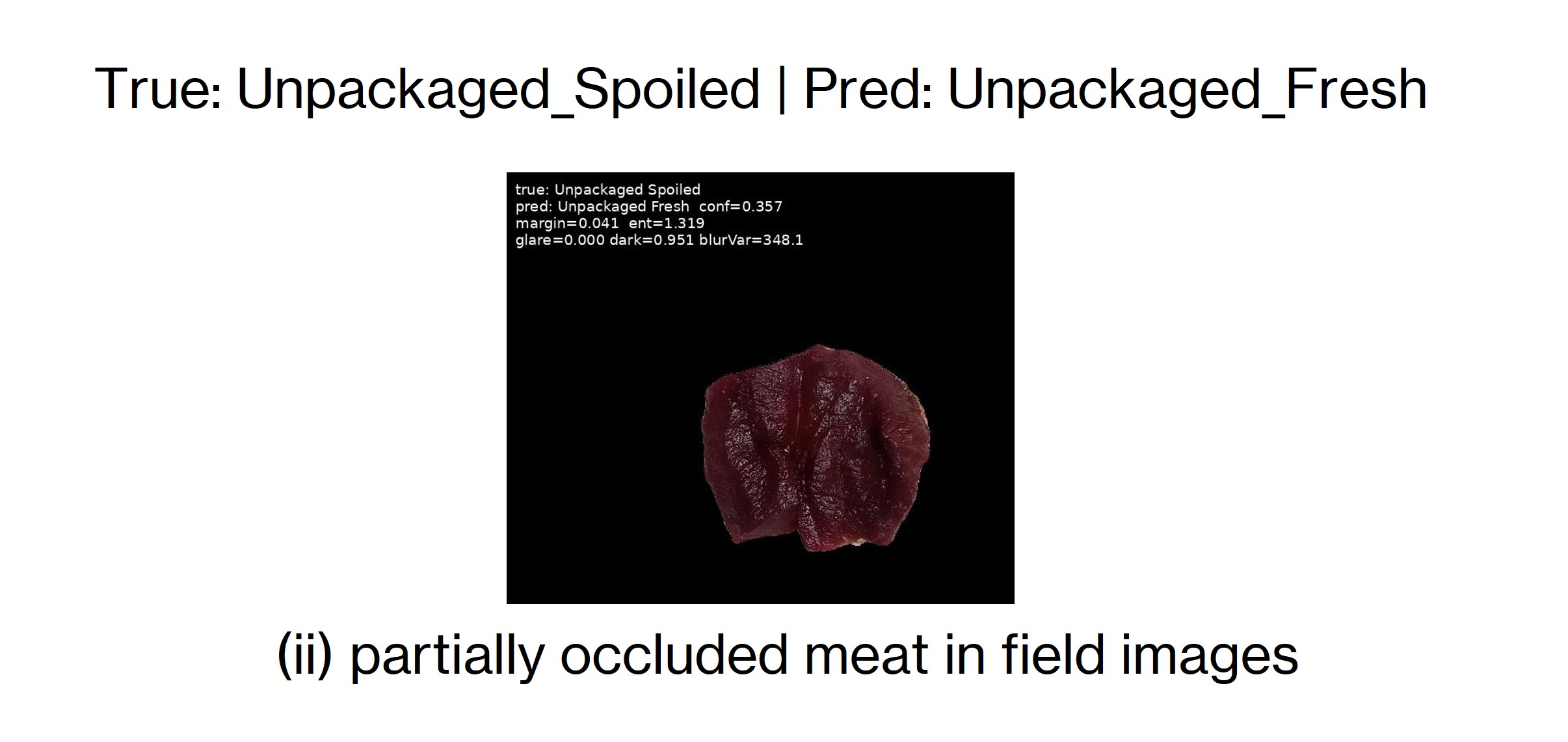}\hfill
\includegraphics[width=0.31\textwidth]{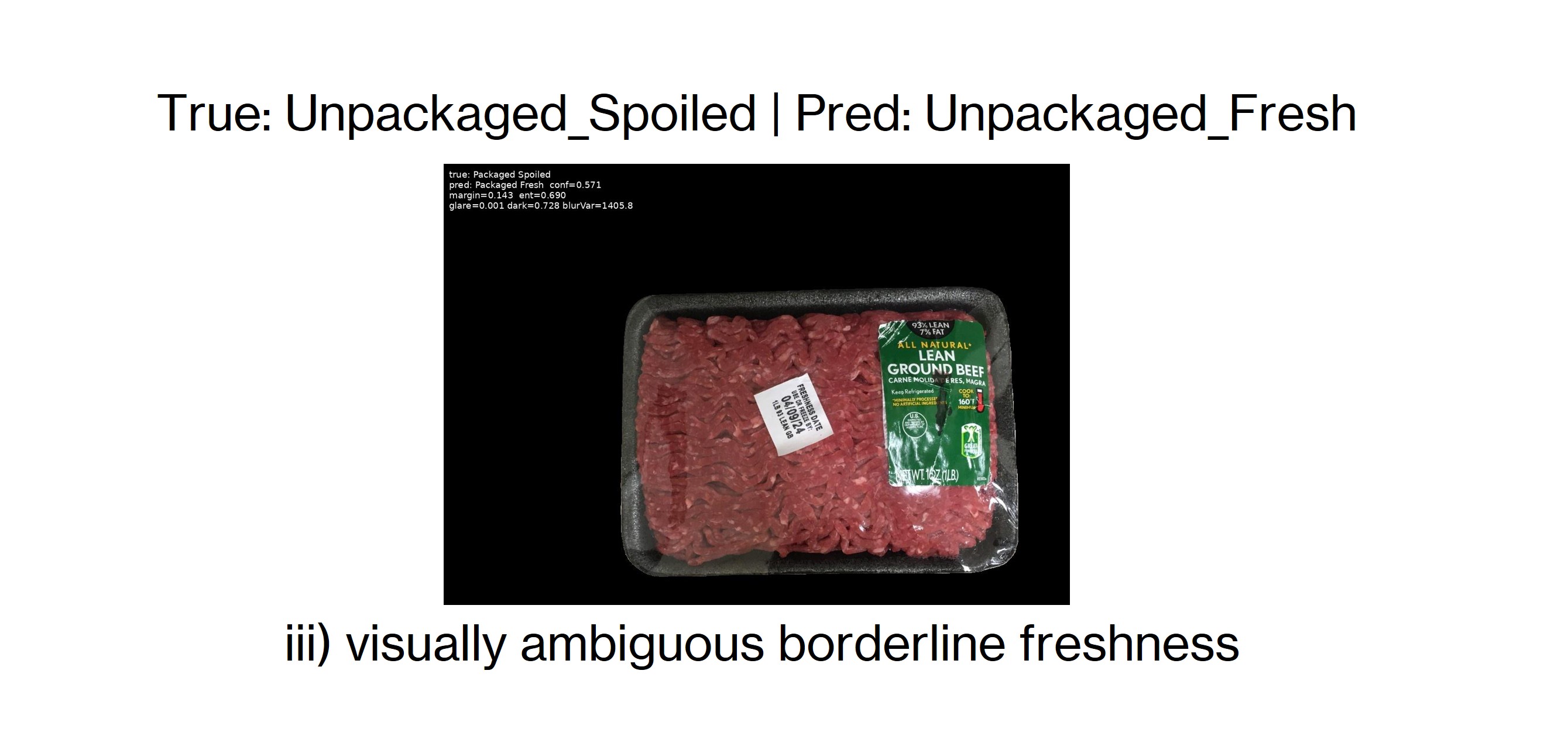}
\caption{Representative failure cases: (i) glare on packaged samples, (ii) occlusion/clutter, and (iii) borderline freshness with subtle cues.}
\label{fig:failure_cases}
\end{figure*}

Figure~\ref{fig:cm_all_models} shows confusion matrices on the held-out ID test set ($N=843$).
Across backbones, misclassifications are concentrated within the same packaging condition (fresh vs.\ spoiled), indicating that the main difficulty lies in subtle freshness cues rather than packaging identification.
For EfficientNet-B0, MobileNetV3-Small, and ResNet-50, the dominant error is Unpackaged Spoiled $\rightarrow$ Unpackaged Fresh (11 images; 4.2\% of the Unpackaged Spoiled class). 
Swin-T exhibits a slightly different pattern, with more Packaged Spoiled $\rightarrow$ Packaged Fresh confusions (10 images; 5.6\%) and a small amount of Unpackaged Fresh $\rightarrow$ Unpackaged Spoiled (8 images; 3.1\%). 
ViT-B/16 shows the largest confusion on the hardest boundaries, especially Unpackaged Spoiled $\rightarrow$ Unpackaged Fresh (30 images; 11.5\%) and Packaged Fresh $\rightarrow$ Packaged Spoiled (13 images; 8.9\%), consistent with its lower macro-F1 in Table~\ref{tab:perclass_models}.
To make the remaining errors more concrete, Fig.~\ref{fig:failure_cases} shows representative failure cases.
These images are challenging because the visual cues are weak or distorted, such as strong glare on packaged trays, partial occlusion/clutter, or borderline freshness where appearance changes are subtle.

\subsection{OOD Detection Performance}
\label{sec:ood_detection}

\begin{table*}[t!]
\centering
\caption{OOD detection results across scoring methods on the OOD benchmark.}
\label{tab:ood_methods}
\renewcommand{\arraystretch}{1.15}
\setlength{\tabcolsep}{6pt}
\begin{tabular}{llccc}
\toprule
\textbf{Model} & \textbf{Method} & \textbf{AUROC} & \textbf{AUPR} & \textbf{FPR@95TPR} \\
\midrule
\multirow{3}{*}{ResNet-50} 
& MSP    & 0.927 & 0.911 & 0.242 \\
& ENERGY & 0.939 & 0.941 & 0.254 \\
& ODIN   & 0.930 & 0.927 & 0.281 \\
\midrule
\multirow{3}{*}{Swin-T}
& MSP    & 0.981 & 0.992 & 0.0250 \\
& Energy & 0.996 & 0.998 & 0.0025 \\
& ODIN   & 0.997 & 0.997 & 0.0025 \\
\midrule
\multirow{3}{*}{ViT-B/16}
& MSP    & 0.993 & 0.947 & 0.0167 \\
& Energy & 0.999 & 0.994 & 0.0083 \\
& ODIN   & 0.998 & 0.988 & 0.0167 \\
\midrule
\multirow{3}{*}{EfficientNet-B0}
& MSP    & 0.952 & 0.954 & 0.206 \\
& ENERGY & 0.982 & 0.977 & 0.083 \\
& ODIN   & 0.977 & 0.972 & 0.103 \\
\midrule
\multirow{3}{*}{MobileNetV3-Small}
& MSP    & 0.971 & 0.978 & 0.240 \\
& ENERGY & 0.978 & 0.982 & 0.131 \\
& ODIN   & 0.985 & 0.987 & 0.106 \\
\bottomrule
\end{tabular}
\end{table*}

We include MSP as a simple baseline confidence score \cite{hendrycks2017}. 
However, softmax confidence can be overconfident even on wrong or OOD inputs, so MSP may fail in practice \cite{guo2017calibration}. 
We therefore also test ODIN, which uses temperature scaling and a small input perturbation to improve OOD separation \cite{liang2018odin}. 
We also test the energy score, which uses logits and is often stronger than softmax-based confidence for OOD detection \cite{liu2020energy}. 
In short, MSP is the lightweight baseline, while ODIN and Energy are stronger options with different runtime and tuning costs \cite{liang2018odin,liu2020energy}.

\begin{figure}[t]
    \centering
    \includegraphics[width=0.9\linewidth]{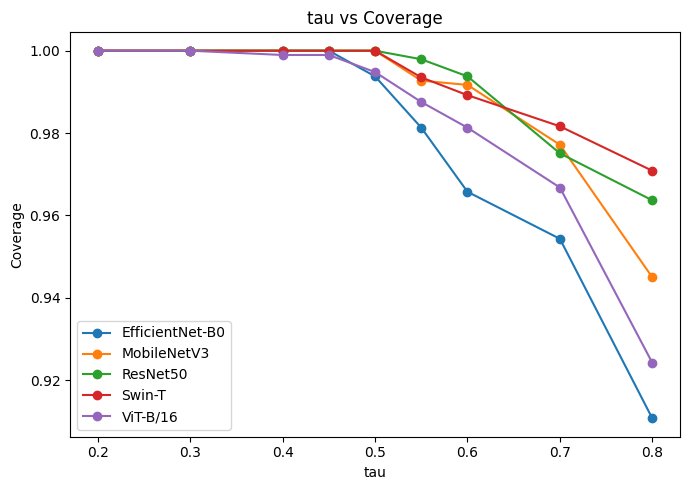}
    \caption{Threshold sweep on the in-distribution (4-class) set: $\tau$ vs.\ coverage (fraction of samples kept).}
    \label{fig:tau_cov}
\end{figure}

\begin{figure}[t]
    \centering
    \includegraphics[width=0.9\linewidth]{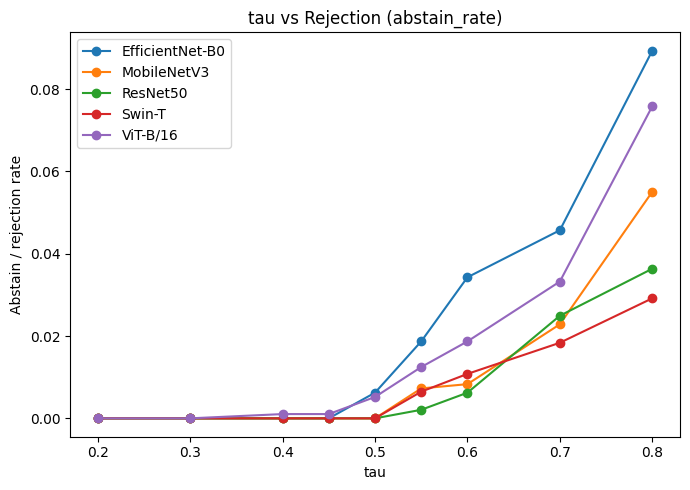}
    \caption{Threshold sweep on the in-distribution (4-class) set: $\tau$ vs.\ rejection rate (fraction of samples abstained).}
    \label{fig:tau_rej}
\end{figure}

While Table~\ref{tab:retrain_all} summarizes ID classification accuracy on the four in-distribution classes, we also measure how well each scoring method separates in-distribution and OOD samples.
Table~\ref{tab:ood_methods} reports AUROC, AUPR, and FPR@95TPR for MSP, Energy, and ODIN.
Across backbones, Energy and ODIN often improve AUROC/AUPR over MSP, although the best method can vary by model. We also perform a sensitivity sweep over $\tau \in \{0.2, 0.3, 0.4, 0.45, 0.5, 0.55, 0.6, 0.7, 0.8\}$ using the MSP confidence score on the in-distribution (4-class) set used for OOD evaluation.
For each $\tau$, we report coverage (fraction of samples kept) and rejection rate (fraction of samples abstained).
Figs.~\ref{fig:tau_cov}--\ref{fig:tau_rej} show that the behavior is stable around $\tau=0.5$, where only a small fraction of the lowest-confidence cases are rejected while coverage remains near-maximal.

In practice, the OOD-aware abstention is most useful as a simple safety check for deployment, especially in consumer-facing scenarios where lighting, glare, and packaging reflections can differ from our training conditions. For that reason, we view the system as a visual screening tool rather than a substitute for microbiological testing or formal inspection. When the input looks atypical or the image quality is poor, a conservative use of the ``No Result'' option can defer the prediction, with such cases left to manual judgment and standard food-safety procedures.

\subsection{Statistical Validation}

\begin{table*}[t!]
\centering
\caption{McNemar paired contingency tables and confidence intervals for pairwise comparison of test accuracy on the held-out ID test set ($N=843$). McNemar $\chi^2$ statistics use the continuity-corrected form $(|n_{10}-n_{01}|-0.5)^2/(n_{10}+n_{01})$ $(\mathrm{df}=1)$.}
\label{tab:mcnemar_ci}
\setlength{\tabcolsep}{6pt}
\renewcommand{\arraystretch}{1.15}
\begin{tabular}{lrrrrrrrr}
\toprule
\textbf{Pair} &
$\mathbf{n_{11}}$ & $\mathbf{n_{10}}$ & $\mathbf{n_{01}}$ & $\mathbf{n_{00}}$ &
$\boldsymbol{\chi^2}$ & $\mathbf{p}$-value &
$\bm{\Delta}\,\mathrm{acc}$ (A--B) & \textbf{95\% CI} \\
\midrule
ResNet50 (A) vs ViT (B) &
788 & 35 & 8 & 12 &
16.331 & $5.32\times10^{-5}$ &
$+0.0320$ & $[0.0169,\ 0.0471]$ \\
ViT (A) vs Swin-T (B) &
778 & 18 & 44 & 3 &
10.512 & $0.0012$ &
$-0.0308$ & $[-0.0490,\ -0.0127]$ \\
ViT (A) vs EfficientNet-B0 (B) &
790 & 6 & 37 & 10 &
21.605 & $3.35\times10^{-6}$ &
$-0.0368$ & $[-0.0518,\ -0.0217]$ \\
ViT (A) vs MobileNetV3-Small (B) &
780 & 16 & 43 & 4 &
11.932 & $0.0005$ &
$-0.0320$ & $[-0.0498,\ -0.0143]$ \\
Swin-T (A) vs MobileNetV3-Small (B) &
815 & 7 & 8 & 13 &
0.0167 & $0.8973$ &
$-0.0012$ & $[-0.0102,\ 0.0078]$ \\
\bottomrule
\end{tabular}
\end{table*}

\begin{table*}[!t]
\centering
\caption{Per-class F1-score with 95\% bootstrap confidence intervals (percentile, $B=4000$) on the held-out ID test set ($N=843$).}
\label{tab:f1_ci}
\renewcommand{\arraystretch}{1.15}
\setlength{\tabcolsep}{4.2pt}
\scriptsize
\begin{tabular}{lccccc}
\hline
\textbf{Class} & \textbf{ResNet-50} & \textbf{Swin-T} & \textbf{ViT-B/16} & \textbf{EfficientNet-B0} & \textbf{MobileNetV3-Small} \\
\hline
Packaged Fresh     & 0.950 (0.921--0.974) & 0.944 (0.916--0.969) & 0.889 (0.847--0.925) & 0.946 (0.918--0.970) & 0.932 (0.899--0.960) \\
Packaged Spoiled   & 0.958 (0.933--0.978) & 0.951 (0.927--0.973) & 0.921 (0.891--0.947) & 0.955 (0.932--0.975) & 0.939 (0.911--0.964) \\
Unpackaged Fresh   & 0.982 (0.970--0.993) & 0.982 (0.970--0.993) & 0.936 (0.913--0.956) & 0.975 (0.960--0.988) & 0.961 (0.943--0.976) \\
Unpackaged Spoiled & 0.983 (0.970--0.993) & 0.983 (0.971--0.993) & 0.928 (0.903--0.951) & 0.974 (0.959--0.987) & 0.958 (0.939--0.975) \\
\hline
Macro Avg          & 0.968 (0.954--0.980) & 0.965 (0.951--0.978) & 0.919 (0.899--0.938) & 0.963 (0.948--0.976) & 0.947 (0.930--0.963) \\
\hline
\end{tabular}
\end{table*}

Table~\ref{tab:mcnemar_ci} reports paired McNemar tests~\cite{japkowicz2011evaluating} on the held-out ID test set ($N=843$), together with 95\% confidence intervals for the paired accuracy difference $\Delta\mathrm{acc}$ (A--B).
Each row includes the full paired contingency table $(n_{11},n_{10},n_{01},n_{00})$, where $n_{10}$ counts test images correctly classified by model A but misclassified by model B (and vice versa for $n_{01}$).
McNemar $\chi^2$ statistics are computed using the continuity-corrected form $(|n_{10}-n_{01}|-0.5)^2/(n_{10}+n_{01})$ (df$=1$), reflecting paired predictions on the same test samples~\cite{yates1934}.

Across comparisons involving ViT-B/16, we observe consistent and statistically significant accuracy gaps in favor of the other backbones.
Specifically, Swin-T outperforms ViT-B/16 by $3.08$ percentage points ($p=0.0012$, $\Delta\mathrm{acc}=-0.0308$, 95\% CI $[-0.0490,-0.0127]$), EfficientNet-B0 by $3.68$ points ($p=3.35\times10^{-6}$, $\Delta\mathrm{acc}=-0.0368$, 95\% CI $[-0.0518,-0.0217]$), and MobileNetV3-Small by $3.20$ points ($p=0.0005$, $\Delta\mathrm{acc}=-0.0320$, 95\% CI $[-0.0498,-0.0143]$).
These intervals do not cross zero, indicating that the observed differences are unlikely to be explained by sampling variability alone.

In contrast, Swin-T and MobileNetV3-Small exhibit nearly identical performance on the ID test set ($\Delta\mathrm{acc}=-0.0012$; 95\% CI $[-0.0102,0.0078]$; $p=0.8973$), indicating no statistically detectable accuracy difference between these two models under paired testing.
Finally, ResNet-50 exceeds ViT-B/16 by $3.20$ percentage points ($p=5.32\times10^{-5}$, $\Delta\mathrm{acc}=+0.0320$, 95\% CI $[0.0169,0.0471]$), consistent with the accuracy ranking in Table~\ref{tab:retrain_all}.

Complementing the accuracy-level hypothesis tests, Table~\ref{tab:f1_ci} summarizes per-class F1-scores with 95\% bootstrap confidence intervals (percentile, $B=4000$).
Overall, the strongest models maintain high and stable F1 across all four ID classes.
The packaged categories show larger uncertainty than the unpackaged classes, which is consistent with additional visual variability introduced by film reflections and glare.
ViT-B/16 yields lower F1 and wider intervals on the harder packaged classes, whereas the CNN backbones and Swin-T retain high F1 with comparatively tighter confidence ranges.

\subsection{Comparison with Prior Works}

\begin{table*}[!t]
\centering
\caption{Quantitative comparison of prior RGB-based meat-freshness studies and this work. Results from this study are reported on the held-out ID test set. OOD performance is evaluated separately (Section~\ref{sec:ood_detection}).}
\label{tab:rgb_comparison}
\renewcommand{\arraystretch}{1.2}
\begin{tabular}{lccc}
\hline
\textbf{Study} & \textbf{Classes} & \textbf{Dataset Size} & \textbf{Reported Metric} \\
\hline
Ulucan et al. (2019)~\cite{ulucan2019} & 2 & 1,896 & 99.6\% (Accuracy) \\
NUS Project (2023)~\cite{bhargav2023meat} & 3 & 2,268 & 93.1\% (Accuracy) \\
Abd~Elfattah et al. (2025)~\cite{abdelfattah2025} & 3 & 2,266 & 98.5\% (Accuracy) \\
\textbf{This Work (2026)} & \textbf{4 (ID)} & \textbf{6,256} & \textbf{98.10\% (EfficientNet-B0), 97.63\% (ResNet-50 / MobileNetV3-Small), 97.51\% (Swin-T)} \\
\hline
\end{tabular}
\end{table*}

Table~\ref{tab:rgb_comparison} places our results alongside prior RGB-based meat freshness studies.
Most earlier work reported high accuracy (93--99\%) but focused on two or three classes and typically used unpackaged samples under controlled acquisition conditions.
In this work, we curate a larger dataset (6,256 images) spanning both packaged and unpackaged meat and report strong held-out ID performance, with EfficientNet-B0 achieving 98.10\% accuracy on the four-class ID test set.

OOD handling is evaluated separately from closed-set classification accuracy.
Rather than treating ``No Result'' as a conventional class, we implement a confidence-based OOD scoring and thresholding to enable abstention on uncertain inputs.
Section~\ref{sec:ood_detection} reports AUROC/AUPR and threshold sweeps that characterize the trade-off between coverage and rejection, demonstrating that the system can defer low-confidence cases instead of forcing overconfident predictions.

Conventional sensing approaches such as spectroscopy, gas sensors, and electronic noses~\cite{Wu2022,id5,Borch1996} can be effective but require dedicated hardware and are harder to deploy at scale.
In contrast, using standard RGB images from commodity cameras or smartphones keeps costs low while maintaining competitive recognition performance on challenging visual conditions (e.g., glare and film reflections in packaged trays).

Overall, we extend prior work in three ways:
(1) a broader formulation covering packaged and unpackaged meat, supported by a larger dataset;
(2) an automated segmentation step that standardizes the region of interest before classification; and
(3) an OOD-aware abstention mechanism to manage uncertainty.
Together, these choices broaden the operating conditions and improve practical reliability.

\begin{table*}[!t]
\centering
\caption{Computational efficiency. Test accuracy and training time were measured on our lab GPU setup (RTX~6000 Ada), while mobile latency was benchmarked by us on a Samsung Galaxy A55 (SM-A556E), Android~16, using the TensorFlow Lite Benchmark APK (this work). Accuracy corresponds to the held-out ID test accuracy (Table~\ref{tab:retrain_all}).}
\label{tab:efficiency_mobile}
\renewcommand{\arraystretch}{1.15}
\begin{tabular}{lccccc}
\hline
\textbf{Model} & \textbf{Params (M)} & \textbf{Latency (ms)} & \textbf{Device / Runtime} & \textbf{Accuracy (\%)} & \textbf{Train Time (min/epoch)} \\
\hline
ResNet\textendash 50           & 25.6        & 76.92  & Galaxy A55, Android 16, TFLite Benchmark & 97.63 & 1.10 \\
EfficientNet\textendash B0     & 5.3         & 17.21  & Galaxy A55, Android 16, TFLite Benchmark & 98.10 & 0.76 \\
MobileNetV3\textendash Small   & 2.5--2.9    & 6.36   & Galaxy A55, Android 16, TFLite Benchmark & 97.63 & 0.74 \\
Swin\textendash T (Tiny)       & 27.5--28    & 108.17 & Galaxy A55, Android 16, TFLite Benchmark & 97.51 & 0.78 \\
ViT\textendash B/16            & 86          & 462.75 & Galaxy A55, Android 16, TFLite Benchmark & 94.42 & 0.84 \\
\hline
\end{tabular}
\end{table*}

\subsection{Computational Efficiency}
\label{sec:efficiency}

Table~\ref{tab:efficiency_mobile} summarizes model size, training cost, and on-device inference latency for deployment-oriented comparison.
Test accuracy and training time were measured in our controlled lab environment using an NVIDIA RTX~6000 Ada (48\,GB) with 64\,GB RAM and an AMD Ryzen Threadripper PRO 7945WX, whereas mobile latency was benchmarked by us on a single smartphone platform (Samsung Galaxy A55, SM-A556E; Android~16) using the TensorFlow Lite Benchmark APK.
We benchmarked on-device latency using float32 TFLite models (i.e., without post-training quantization such as FP16 or INT8).
Because all latency results were obtained on the same device and runtime, the reported latencies are directly comparable across models under a consistent measurement protocol.

Among CNN backbones, EfficientNet\textendash B0 and MobileNetV3\textendash Small provide strong parameter efficiency (5.3\,M and 2.5--2.9\,M parameters) while maintaining high held-out ID accuracy (98.10\% and 97.63\%).
EfficientNet\textendash B0 offers the best accuracy--latency balance in our setting: it attains the highest ID test accuracy (98.10\%) while remaining fast on-device (17.21\,ms), substantially lower than ResNet\textendash 50 (76.92\,ms) and Swin\textendash T (108.17\,ms).
MobileNetV3\textendash Small is the fastest model (6.36\,ms) with competitive accuracy (97.63\%), making it attractive under tight real-time constraints.
ResNet\textendash 50 achieves similar accuracy (97.63\%) but at a larger parameter budget (25.6\,M) and higher mobile latency.

For Transformer-based models, Swin\textendash T reaches competitive accuracy (97.51\%) with a parameter count comparable to ResNet\textendash 50 (27.5--28\,M), but incurs higher on-device latency (108.17\,ms).
ViT\textendash B/16 shows the least favorable trade-off in our setting, combining the largest footprint (86\,M) with the highest latency (462.75\,ms) and the lowest ID accuracy (94.42\%).

Training time per epoch varies only modestly across models (0.74--1.10 minutes/epoch), indicating that deployment-time inference latency is the primary differentiator for edge use.
Overall, the results suggest that lightweight CNNs are the most practical option for real-time mobile inference under tight compute budgets, while higher-capacity backbones such as ResNet\textendash 50 or Swin\textendash T may be preferred when model capacity is prioritized and higher latency is acceptable.
Future work will standardize additional inference settings (e.g., quantization level and thread configuration) to quantify the impact of hardware-aware optimization on mobile performance.

\section{Conclusion}
\label{sec:conclusion}

In this work, we proposed a meat-freshness assessment framework from RGB images that covers both packaged and unpackaged meat and incorporates an explicit \emph{No Result} option to defer ambiguous inputs via OOD-aware abstention.
The pipeline integrates U-Net segmentation with deep classifiers and applies nested five-by-three cross-validation for model selection and unbiased evaluation.
The segmentation stage provided consistent region localization (mean IoU \(\approx 0.75\), mean Dice \(\approx 0.82\)), helping standardize inputs and stabilize training.

On the held-out ID test set (\(N=843\)) with four meat classes, EfficientNet-B0 achieved the highest accuracy (98.10\%), followed by ResNet-50 and MobileNetV3-Small (both 97.63\%), and Swin-T (97.51\%), while ViT-B/16 was lower (94.42\%).
Per-class and statistical analyses further confirmed that most remaining errors occur within the same packaging condition (fresh vs.\ spoiled), reflecting the subtle visual boundary, especially for packaged samples affected by glare and film reflections.
In addition to closed-set accuracy, we evaluated OOD scoring methods (MSP, Energy, and ODIN) and a threshold sweep to characterize the coverage--rejection trade-off, showing that abstention can reduce overconfident decisions by filtering the lowest-confidence cases.

Finally, deployment-oriented benchmarking indicates that lightweight CNNs provide the best accuracy--latency balance on mobile hardware: EfficientNet-B0 and MobileNetV3-Small retain high ID accuracy with substantially lower on-device latency than larger backbones.
Overall, combining segmentation, OOD-aware inference, and nested cross-validation yields a reliable and scalable baseline for practical meat freshness detection.

The experiments were carried out in a controlled setting to ensure repeatability and establish a clear benchmark.
Future work may expand data collection to cover a wider range of lighting conditions, camera types, and spoilage stages, supported by expert annotation and laboratory reference measures. A focused quantitative ablation that compares classification performance with and without the segmentation step (under the same backbone and evaluation protocol) would help isolate the contribution of foreground extraction. External validation on a fully independent dataset, even at small scale, would further test robustness to new acquisition conditions and product variations. To strengthen labeling credibility, a small expert-reviewed subset could be used to report inter-annotator agreement (or reviewer consistency) and to benchmark how closely the current labels align with domain judgment.

Microbiological and physicochemical validation could be incorporated using pH, psychrotroph counts, and standard microbial assays such as \textit{E.~coli} and \textit{Salmonella} detection. In addition, short image sequences can be evaluated for temporal modeling of spoilage progression, explainability tools (Grad-CAM and LIME) may be integrated with expert feedback, and multimodal sensing that combines RGB with spectroscopy or gas-sensor inputs can be explored. These directions are expected to move the framework toward a more interpretable and practically deployable system for real-world food safety monitoring.

\section*{Acknowledgment}

We appreciate Colton Bailey and Josephine Nyame's help in the data collection process for this project. 
We also thank Yuchen Tian, Sam Chamoun, and Andrea Panebianco for their valuable discussions, feedback, and suggestions during the development and revision of this manuscript. 
Their insights contributed to improving the clarity and technical accuracy of the paper.

\begin{IEEEbiography}[{\includegraphics[width=1in,height=1.25in,clip,keepaspectratio]{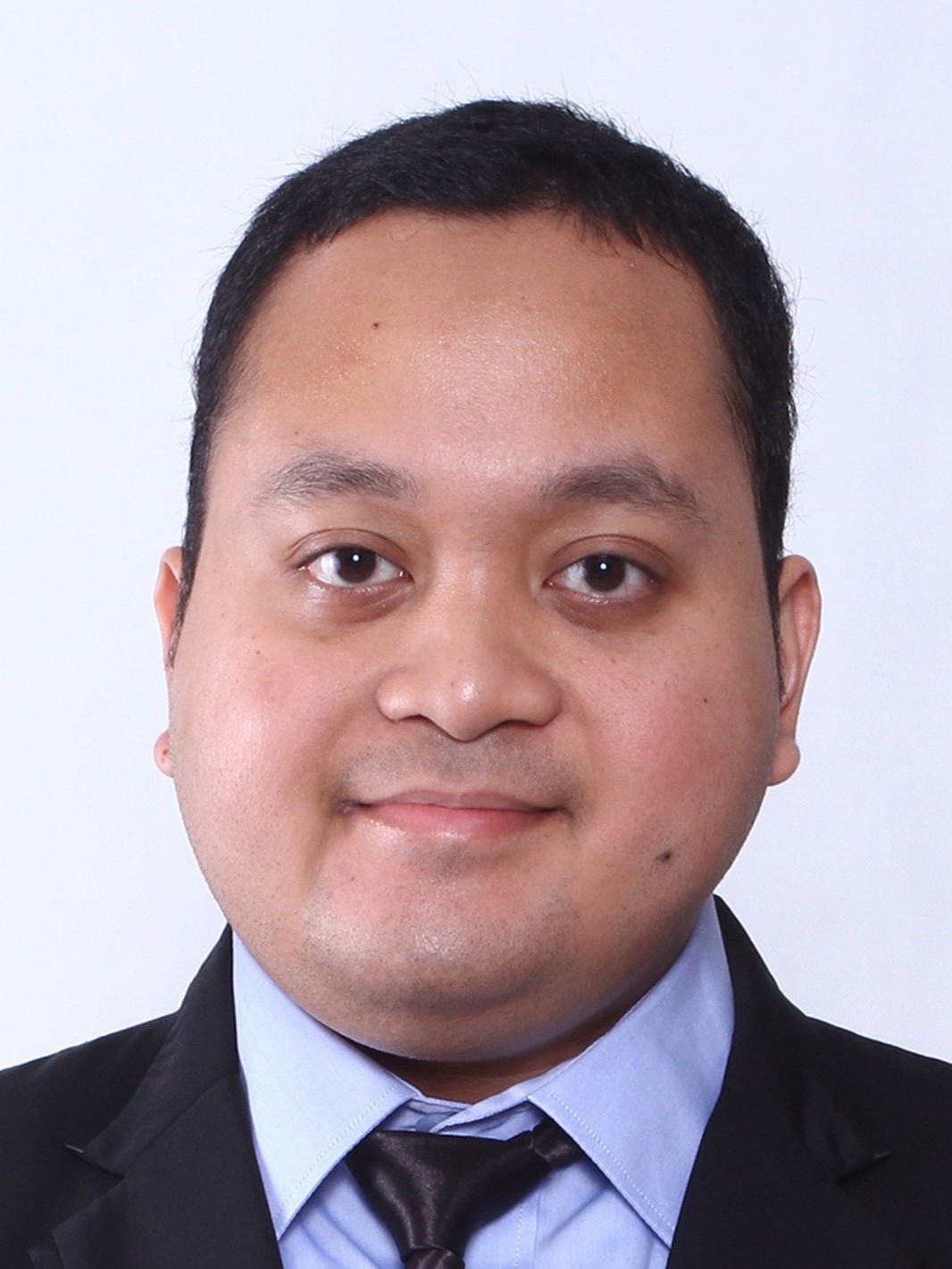}}]{Hutama Arif Bramantyo} is currently pursuing the Ph.D. degree in Electrical and Computer Engineering at Auburn University, AL, USA, as a Fulbright Doctoral Degree Grantee. He is also a Lecturer in the Department of Telecommunication Engineering at Politeknik Negeri Semarang, Indonesia.
He received the B.Eng. degree in Electrical Engineering from Diponegoro University, Indonesia, in 2014, and the M.Eng. degree in Telecommunication Management from the University of Indonesia in 2019. Prior to academia, he worked in the telecommunications industry for over six years at Lintasarta, where he held various engineering and planning roles.
His research interests include wireless communications, Internet of Things (IoT), antenna design, and the application of machine learning in agriculture and food systems. Mr. Bramantyo has published in IEEE conferences and national journals and has contributed to several community service and technology transfer initiatives related to digital agriculture and ICT for rural development.
\end{IEEEbiography}
\vspace{-1.2\baselineskip}
\begin{IEEEbiography}[{\includegraphics[width=1in,height=1.25in,clip,keepaspectratio]{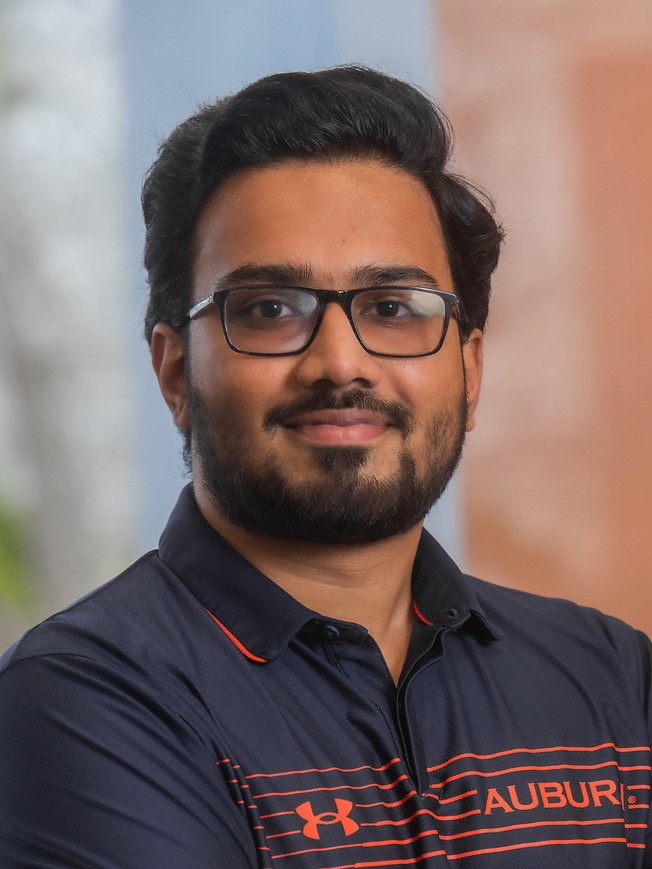}}]{Mukarram Ali Faridi} (Student Member, IEEE) received the B.Eng. degree in Computer Engineering from Auburn University, Auburn, AL, USA, in 2022 and the M.S. degree in Electrical Engineering from the same institution in 2025, where his graduate studies centered on VLSI testing, chip architecture and fabrication, and machine learning.
He is currently an FPGA Engineer with Adtran, Richardson, TX, USA. Prior to joining industry, he was a Graduate Research Assistant at Auburn University, leading an NSF- and OCP-supported investigation into silent data corruption in integrated circuits, and an Embedded Software Engineer Intern at Club Car, Sarasota, FL, USA, where he developed an ESP32-based vehicle simulation automation platform. His research interests include VLSI testing, embedded systems, and real‑world applications of machine learning, particularly in agriculture and food systems.
Mr. Faridi received the Auburn University Woltosz Fellowship and a full‑tuition scholarship for his graduate studies. As an undergraduate, he graduated \textit{magna cum laude}, earned University Honors Scholar distinction, and was recognized among the top 20 undergraduates in the College of Electrical and Computer Engineering. He has published his VLSI testing research in IEEE conference proceedings.
\end{IEEEbiography}
\vspace{-1.2\baselineskip}

\begin{IEEEbiography}[{\includegraphics[width=1in,height=1.25in,clip,keepaspectratio]{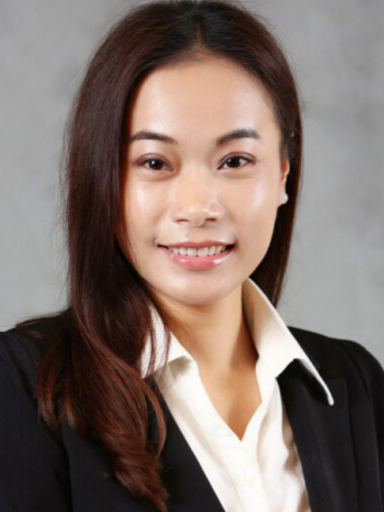}}]{Rui Chen} is an Assistant Professor in the Department of Agricultural and Environmental Sciences at Tuskegee University, AL, USA. She serves as an agricultural economist in the farmer technology adoption decisions thrust of the Artificial Intelligence for Future Agricultural Resilience, Management, and Sustainability (AIFARMS) Institute. 
She received the M.A. degree in Applied Economics from Ocean University of China in 2013 and earned the Ph.D. degree in Applied Economics, with a minor in Statistics, from Auburn University in 2017. From 2017 to 2020, she was a Postdoctoral Researcher and Instructor at Auburn University. Her current research interests include experimental and food economics, agricultural technology adoption, and the application of machine learning in agricultural and food systems. 
Dr. Chen has received several honors, including the New Investigator Award from USDA-NIFA (2023), the Outstanding Young Professional Award from the Committee on the Opportunities and Status of Blacks in Agricultural Economics (COSBAE) and the Committee on Women in Agricultural Economics (CWAE) of the Agricultural and Applied Economics Association (AAEA) (2022), and the EXCEL Faculty Fellowship from the 1890 Universities Foundation (2023). She secured millions of dollars in funding from the USDA to support her research in agricultural economics, machine learning, and public policy.
\end{IEEEbiography}
\vspace{-1.2\baselineskip}
\begin{IEEEbiography}[{\includegraphics[width=1in,height=1.25in,clip,keepaspectratio]{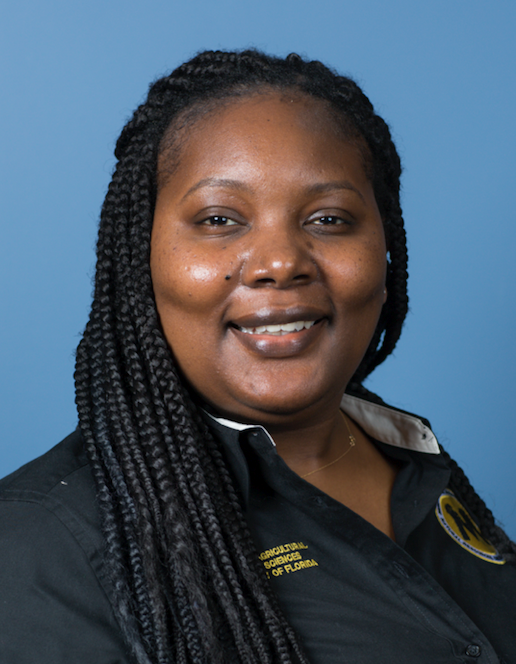}}]{Clarissa Harris} received the B.S. and M.S. degrees in Biology and Biotechnology from Fort Valley State University, Fort Valley, GA, USA, in 2010 and 2013, respectively, and the Ph.D. degree in Animal Science from the University of Florida, Gainesville, FL, USA, in 2020. Her doctoral research focused on the antimicrobial effects of vinegar-based ingredients and lauric arginate against \textit{Salmonella }on poultry products.
She is currently a Research Extension Assistant Professor at Tuskegee University, Tuskegee, AL, USA. Her research interests include food safety, microbiological decontamination of meat, poultry processing, and antimicrobial intervention technologies.
Dr. Harris has authored publications in journals such as The Journal of Applied Poultry Research and Direct Research Journal of Agriculture and Food Science. She has presented her work at numerous national conferences including the Poultry Science Association and MANRRS. She is a recipient of multiple awards, including the Booker T. Washington Leadership Institute Grant and the Delores A. Auzenne Dissertation Award. She was also selected as a 2025 Emerging Leader through the meat Institute. She holds certifications in HACCP and Preventive Controls for Animal Food, and is actively involved in outreach and mentoring through organizations such as MANRRS and LSAMP.
\end{IEEEbiography}

\begin{IEEEbiography}[{\includegraphics[width=1in,height=1.25in,clip,keepaspectratio]{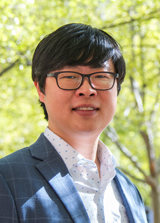}}]{Yin Sun} (Senior Member, IEEE) is the Bryghte D. and Patricia M. Godbold Endowed Associate Professor in the Department of Electrical and Computer Engineering at Auburn University, Alabama. He received his B.Eng. and Ph.D. degrees in Electronic Engineering from Tsinghua University in 2006 and 2011, respectively.
From 2011 to 2017, he was a Postdoctoral Scholar and Research Associate at The Ohio State University. He joined Auburn University in 2017 as an Assistant Professor and was promoted to Associate Professor in 2023. His research interests include Semantic and Goal-oriented Communications, Wireless Networks, and Applied Artificial Intelligence in Agriculture. Dr. Sun has served on the editorial boards of the \emph{IEEE/ACM Transactions on Networking}, \emph{IEEE Transactions on Information Theory}, \emph{IEEE Transactions on Network Science and Engineering}, \emph{IEEE Transactions on Green Communications and Networking}, and the \emph{Journal of Communications and Networks}. He has also served on the organizing committees of numerous international conferences, including as Technical Program Committee Chair for ACM MobiHoc 2025 and General Chair for IEEE/IFIP WiOpt 2026. He founded the Age of Information (AoI) Workshop in 2018 and the Modeling and Optimization in Semantic Communications (MOSC) Workshop in 2023. His publications have received multiple recognitions, including the Best Student Paper Award at IEEE/IFIP WiOpt 2013, the Best Paper Award at IEEE/IFIP WiOpt 2019, runner-up for the Best Paper Award at ACM MobiHoc 2020, the Best Paper Award from the \emph{Journal of Communications and Networks} in 2021, and the IEEE Communications Society William R. Bennett Prize in 2025. He received the Auburn Author Award in 2020 and the National Science Foundation (NSF) CAREER Award in 2023. 
\end{IEEEbiography}

\EOD

\end{document}